\documentclass[journal]{IEEEtran}
\usepackage{graphicx}
\usepackage{subcaption}
\usepackage{array}
\usepackage{amsmath}
\usepackage{algorithm}
\usepackage{algorithmic}
\usepackage{color}
\usepackage{tabularx,multirow}
\usepackage{makecell}

\usepackage{cite}

\begin{document}

\title{Tensor-based Cooperative Control for Large Scale Multi-intersection Traffic Signal Using Deep Reinforcement Learning and Imitation Learning}
%

\author{Yusen~Huo,
        Qinghua~Tao, ~
        and Jianming~Hu

\thanks{
\emph{(Corresponding author:Jianming Hu)}}
\thanks{Yusen Huo is with the Department
of Automation, Tsinghua University, Beijing,
 100084 China, E-mail: huoyusen1@gmail.com }
\thanks{Qinghua Tao is with the Department
of Automation, Tsinghua University, Beijing,
 100084 China, E-mail: taoqh14@mails.tsinghua.edu.cn  }
\thanks{Jianming Hu is with the Department
of Automation, Tsinghua University, Beijing,
 100084 China E-mail:hujm@mail.tsinghua.edu.cn}
}

\markboth{}%
{Shell \MakeLowercase{\textit{et al.}}: Bare Demo of IEEEtran.cls for IEEE Journals}

\maketitle

\begin{abstract}
Traffic signal control has long been considered as a  critical topic in intelligent transportation systems.  Most existing learning methods mainly focus on isolated intersections and suffer from inefficient training. This paper aims at the cooperative control for large scale multi-intersection traffic signal, in which  a novel end-to-end learning based model is established and the efficient training method is proposed correspondingly. In the proposed model, the input traffic status in multi-intersections is represented by a tensor, which not only significantly reduces dimensionality than using a single matrix but also avoids information loss. For the output,
a multidimensional boolean vector is employed for the control policy to indicate whether the signal state changes or not, which simplifies the representation and abides the practical phase changing rules. In the proposed model, a multi-task learning structure is used to get the cooperative policy by learning.
Instead of only using the reinforcement learning to train the model, we employ imitation learning to integrate a rule based model with neural networks to do the pre-training, which provides a reliable and satisfactory stage solution and greatly accelerates the convergence. Afterwards, the reinforcement learning method is adopted to continue the fine training, where proximal policy optimization algorithm is incorporated to solve the policy collapse problem in multi-dimensional output situation. In numerical experiments,  the advantages of the proposed model are demonstrated with comparison to the related state-of-the-art  methods.
\end{abstract}

\begin{IEEEkeywords}
Tensor, multi-intersection, deep reinforcement learning, imitation learning, proximal policy optimization.
\end{IEEEkeywords}

\IEEEpeerreviewmaketitle

\section{Introduction}\label{intro}

\IEEEPARstart{T}{ransportation}  systems play increasingly important roles in modern society \cite{POUMANYVONG2012268,Pojani_2015}. Among the key factors influencing the traffic networks, signal control on intersections plays a vital role in the operation efficiency of the  system. Thanks to the development of intelligent computing technologies \cite{bitam2012its,7564647}, the real-time central control systems for traffic networks  become more feasible. The vehicle-to-vehicle (V2V) and vehicle-to-infrastructure (V2I)  communications provide new approaches for traffic signal control at intersections \cite{liu2013opportunities,7992934}, in which the signal controllers can obtain  accurate and dynamic information of the vehicles in real time to adaptively optimize the system. Therefore, efficient and accurate traffic signal control attracts wider attention in practice.

There are mainly two  approaches for the signal control  in intersections. One is the rule-based approach, whose main idea is to transform the control problem into an operational programming problem, such as  cell transmission model (CTM)\cite{timotheou2015distributed,8270580}. Although this kind of methods have  analyzable  results, they suffer from drastic computational burden when the environment complexity increases, and substantial simplification have to be made to enable the problem tractable, which ultimately affects the performance. The other kind is based on machine learning 
\cite{CASTRO20171182,7424092,li2016traffic,genders2016using}.
Compared with the rule-based approach, these methods are more advantageous in dealing with complex scenarios.

In fact, the traffic signal controls can be regarded as typical sequence decision problems, which well fit into the reinforcement learning (RL) framework \cite{sutton2018reinforcement}. Many researchers have been trying to use RL methods to solve the problem from different perspectives, and have shown better performance than other learning-based methods \cite{huo2018a,el2013multiagent,liang2018deep,mannion2016experimental,Zeng2018AdaptiveTS}. The early proposed RL based models mainly used the Q-learning method  to convert the traffic states to discrete values, but they suffers from the curse of dimensionality due to the storage of large Q-table \cite{el2013multiagent,mannion2016experimental}. Therefore, it often makes a rough division of traffic states at the cost of information loss.

With the development of deep learning, the value function evaluation methods based on deep neural networks are able to conquer the problem of dimensionality curse in RL methods \cite{mnih2013playing}. Li et al. \cite{li2016traffic} and Noe \cite{Casas2017Deep} used the real-time human feature as input states to expand the applicability of the model to a certain extent. However, these methods cannot completely avoid the problem of information loss. In \cite{genders2016using}, it transformed real-time traffic states into position matrices and speed matrices, and adopted convolutional networks to automatically extract features \cite{genders2016using}. This method solves the problem of information loss of input states, but it only models a single intersection and does not solve the problems of spatial coupling. Then Liu et al. tackled the state sharing problem of multiple intersections by concatenating the position matrices of multiple intersections \cite{Liu2017CooperativeDR}. However, they concatenated the matrices just by simple splicing, which greatly increases the dimension of the inputs and is not easy to extend. In addition, most existing deep RL based models for traffic signal control transform the input states into the values of different actions where the highest values are used as the actions for each decision \cite{li2016traffic,genders2016using,Liu2017CooperativeDR,liang2018deep,van2016deep, Zeng2018AdaptiveTS}. However, such models can only deal with the isolated intersections, and lack extendability.

This paper establishes a new end-to-end cooperative control model for large scale multi-intersection traffic signal, which is named as deep reinforcement learning and imitation learning based  multi-intersection signal control model (DRI-MSC). 
Instead of using a huge matrix to cover all the traffic information of the multi-intersections located in the selected area,  the proposed DRI-MSC model constructs a tensor to represent the traffic information, where each matrix describes one intersection, which reduces dimensionality  with computational intractability and avoids information loss. Unlike the four-phase output, we convert the output of the multi-intersection into a multi-dimensional boolean vector, where the real phase changing rules are considered to simulate the practical scenarios and simplify the representation. Moreover,  the multi-task learning method is employed to solve the cooperative control problem in multi-intersections, which helps avoid the local optima caused by rule based cooperative policy. The proposed model has simpler structure together with better extendability, scalability and efficiency to large scale problems.

In the training part,  the proposed DRI-MSC model first introduces the supervised imitation learning method to initialize the training with a satisfactory solution, which greatly accelerate the poor convergence in deep RL based methods. Then, RL method is employed to fine-tune the model. The combined training method relieves a large number of invalid searches to greatly improve the convergence rate and also provides more possibilities to skip the local optima. In addition, the proposed method also  alleviates the oscillation amplitude of the policy in convergence. Finally,  for the policy collapse problem in multi-dimensional output situation, we employ the proximal policy optimization (PPO) algorithm to reduce the impact of policy collapse to a large extent via clipping policies  \cite{zhang2018efficient,2018-TOG-deepMimic,OpenAI_dota}.

The remainder of the paper is arranged as follows. Section \ref{sect1} introduces the background of modeling and training. Section \ref{sect2} presents details of the proposed DRI-MSC model. Numerical experiments based on different traffic flows are conducted in Section \ref{sect3}, and Section \ref{sect4}  ends the paper with conclusion.

\section{Background}\label{sect1}

\subsection{Traffic Signal Control Model}
\begin{figure}[ht]
\centering
\includegraphics[width=0.36\linewidth]{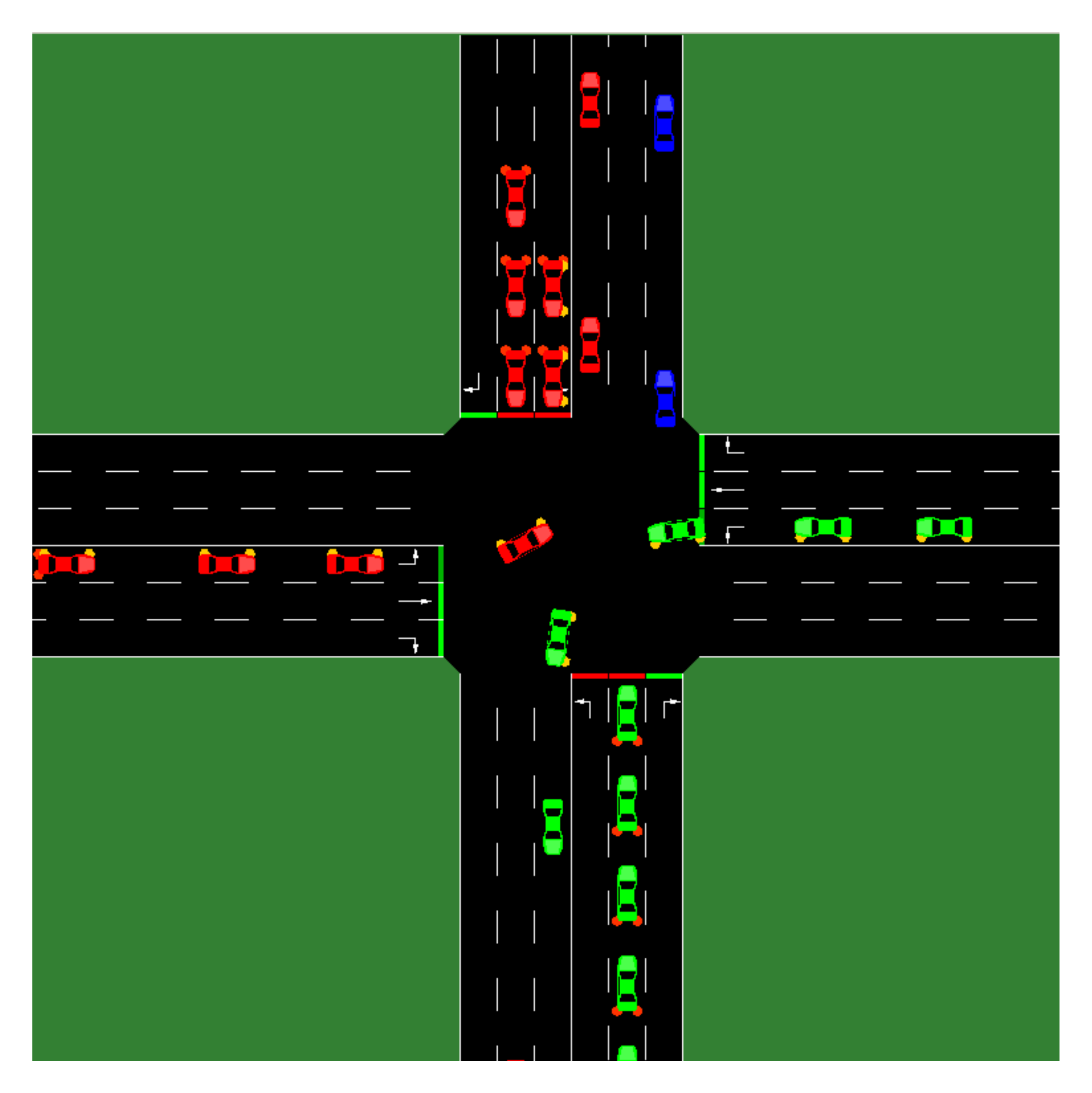}
\includegraphics[width=0.4\linewidth]{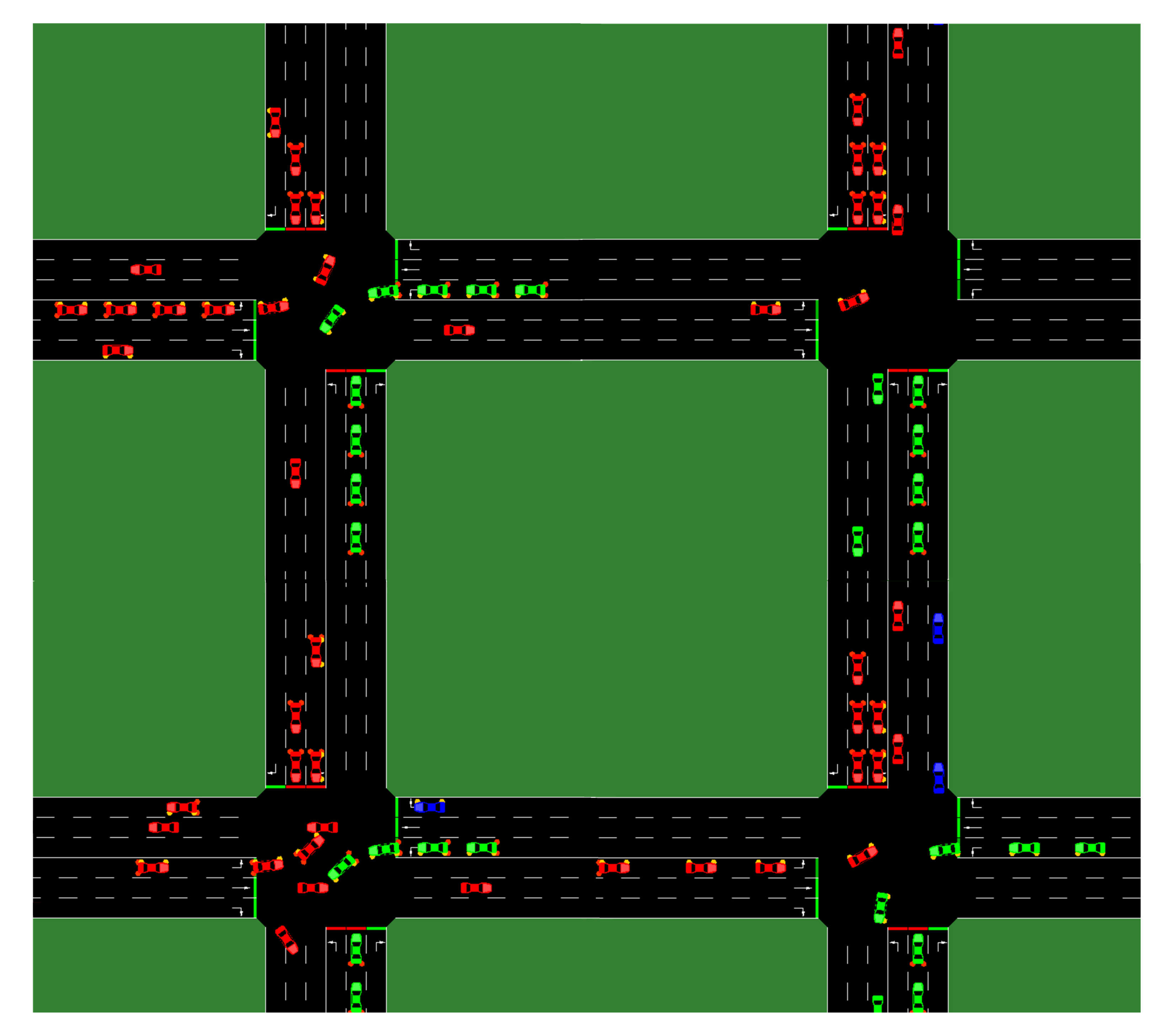}
\caption{The left sub-figure shows a traffic network with single intersection, while the right sub-figure with 4 intersections.}
\label{intersection}
\end{figure}

Intersections are considered as the bottlenecks of traffic network, where vehicles gather and dissipate. The most common intersections are the  4-way and 3-way (T-junction) type cross road. In the modern traffic systems, there are  multiple intersections in nearby areas, which coordinate to control the traffic networks. Fig. \ref{intersection} shows the traffic networks with single intersection and 4 intersections.

In order to model the multi-intersection signals, the input features, the model output, the collaborative control policy and the evaluation indexes should be taken into account. Typically, features such as traffic flow size, average vehicle speed and lane occupancy are often used to reflect traffic states. Generally, these features are represented as a matrix, such as position matrix for specific description to the traffic states \cite{genders2016using}, which inspire the model input of this paper.

\begin{figure}[ht]
\centering
\includegraphics[width=0.36\linewidth]{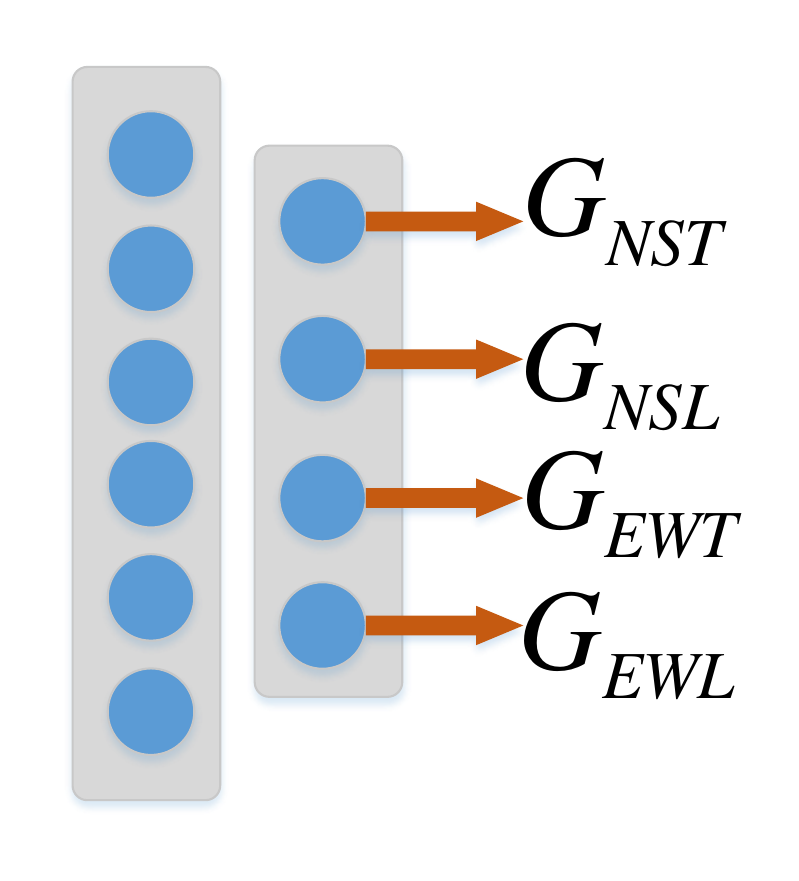}
\includegraphics[width=0.36\linewidth]{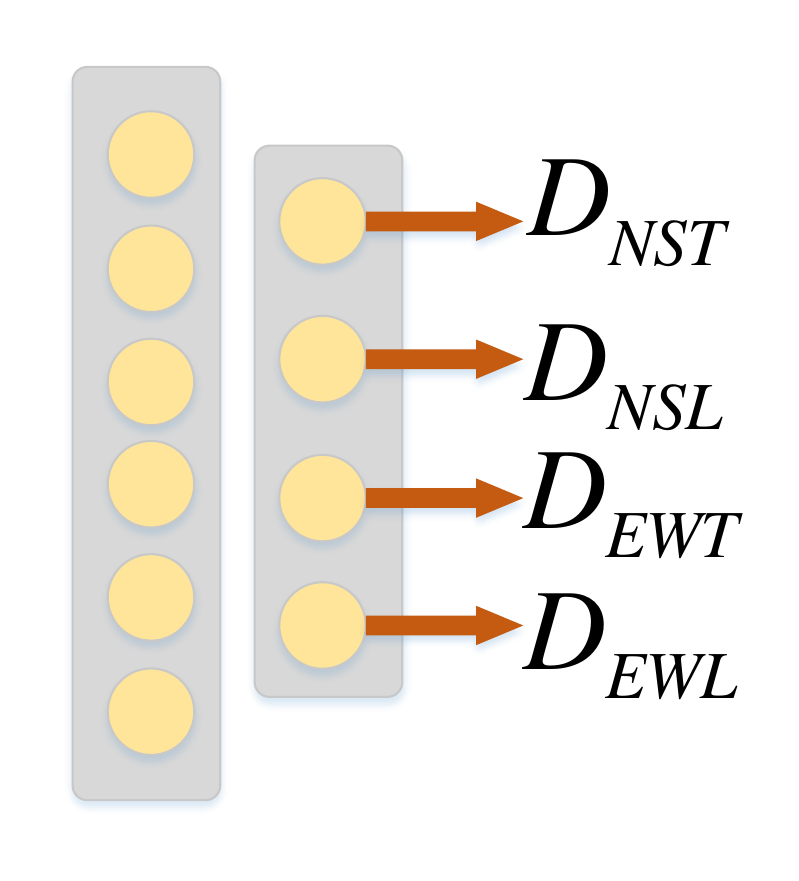}
\caption{The left is for the phase output,  and the right for the phase duration, where  ``G' and ``D'' represent the decision and duration of each green phase respectively. In the first two digits in the subscript,  ``N'',  ``S",  ``E"  and  ``W"  represent  the  north,  south,  east  and west respectively,  while the last digit ``T" and ``L" denote the through and left-turn direction. }
\label{output}
\end{figure}

The output of the control model can be converted to different phases of traffic lights, which control the traffic flow directly. Typically,  control model either output the duration of green phases for different directions, or just output the decision whether the green phase change to the directions\cite{liang2018deep,Casas2017Deep,genders2016using}. We name them as the phase based output. For the former one which is shown in the left of Fig. \ref{output}, the sequence of phases is disrupted which is forbidden in the real world. For the latter one shown in the left, the  sequence of phases can be guaranteed at the cost of the hysteresis of the timing policy.

 Typically, In order to output collaborative control policy of multi-intersections, human-designed rules are used to modify the output of sub-agents to get the cooperative output, which is difficult to design and prone to fall into the local optima.

For the objective, indexes such as queue length, waiting time of vehicles, fuel consumption and accumulative delay are often used to quantify the traffic efficiency. Then programming methods and learning-based methods are used to calculate the optimal value of the these indexes. For learning-based methods, which are applied in this paper, RL are mostly used, and the indexes are converted into rewards and maximized through learning process.

\subsection{Reinforcement learning}
RL is an adaptive learning framework whose idea comes from biological mechanism, and it is a method for agents to constantly change and optimize their policies based on the past experience \cite{sutton2018reinforcement}. The main characteristic of RL is to learn good policies through exploration by trial-and-error and exploitation by the existing experience. In the decision-making process, RL based models map each state $s_t$ to an action $a_t$, and then receive rewards from environmental feedback to continuously optimize its policies accordingly. The  goal of RL is to maximize the  total future reward, which can be measured with a value function $v_{\pi }(s_t)$, regarding the cumulative expectation of the rewards under policy $\pi$, i.e.,
\begin{equation}\label{value_equ}
v_{\pi }(s_t)=E_{\tau \sim \pi}\left [ \sum_{k=t}^{T}\gamma ^{k}R_k \right ],
\end{equation}
where $t$ represents the current time-step, $\tau$ denotes the trajectory history, $R_k$ is the reward of one time-step $k$, and $T$ means the terminal time-step. In addition,  the future reward at each time-step $k$ can have a decay with a certain rate $ 0< \gamma <1$, meaning that the states far away from the current time-step have lower weights. For the traffic control problem considered in this paper, larger $\gamma$ gives higher weight to the long-term traffic efficiency.

RL methods can be mainly categorized into two types. One is the value-based approach, which makes decisions by calculating the values of different actions in the current state. Deep Q net (DQN) is the typical instance for this idea. As is mentioned in Section \ref{intro}, this method is not suitable for the multi-intersection scenes. Another type is the policy-based approach that maps states to policies, and one representative method is the policy gradient algorithm, which uses a neural network to straightforwardly output multi-dimensional actions \cite{Sutton:1999:PGM:3009657.3009806}. The latter is suitable to control multiple signal controllers simultaneously and more feasible for large scale problems.

\subsubsection{Policy gradient and optimization}
For a policy-based model, the actions are mapped from the parameter space, and the environment then maps the actions to the rewards. Combined with (\ref{value_equ}), we can bridge a functional relationship between parameters $\theta$ and value function $v_{\pi }(s_t)$. The gradient of values with respect to $\theta$  can be calculated by back-propagation algorithm, and then use gradient descent method to update the parameters \cite{Rumelhart1986LearningRB}. The policy gradient $\bigtriangledown_{\theta}v_{\pi }(s_t)$ can be calculated as
\begin{equation}\label{policy_formula}
\bigtriangledown_{\theta}v_{\pi }(s_t)=E_{s_t\sim d^{\pi},a_t\sim\pi}[\bigtriangledown_{\theta}\log\pi_\theta(a_t\mid s_t)A^{\pi}(s_t,a_t)]
\end{equation}
where $d^{\pi}$ denotes the distribution of the $s_t$ under policy $\pi$, and $A^{\pi}(s_t,a_t)$  is called the advantage function \cite{374604}, which represents the additional value under action $a$ compared with the expected value of $s_t$.

\subsubsection{Actor critic}
Practically, $A^\pi(s_t,a_t)$ is hard to  calculate directly, and there are mainly two methods to estimate $A^\pi(s_t,a_t)$, in which the value approximation method  is proved to be better than only using the trajectory data \cite{Konda1999ActorcriticA}. The method  which combines policy gradient with value approximator is called actor critic. In this paper, actor critic is applied to improve the convergence rate of the model, where a neural network is used as an approximator and is called value network.

\subsection{Imitation learning}\label{sectionil}

Imitation learning utilizes models to imitate behaviors of the experts or other models. When the model can accurately predict the behaviors, we only need to make decisions based on the results predicted by such model. 
Moreover, imitation accuracy can also be used to judge the feature extraction and model ability, which means low accuracy may be caused by defective feature extraction method or insufficient model structures. 

A typical method of imitation learning is to train classifiers or regressors to predict the behaviors of the experts based on input states and the  observed output behaviors. However, the behaviors of learners also affect the future input observations and states during the implementation of policies, which results in the violation of the I.I.D. ( independent and identically distributed) hypothesis of the methods. In practice, once a learner makes a mistake, it can encounter observation data which are completely different from the presentation of experts, leading to the compound of errors.

In order to solve this problem, trajectory data should also be added to the training data dynamically. Dataset Aggregation (DAGGER)  collects trajectory data at the same time when model makes decisions, and label the new collected data by experts in real time \cite{Ross2011ARO}. Then the training set can cover most of possible scenes  through repeated training and decision making process, and the imitation learning based model can perform as an expert. In this paper, we use DAGGER to pre-train our model with a simplified rule based model as labels.

 \section{The DRI-MSC model}\label{sect2}
 In this section, we introduce the proposed DRI-MSC model regarding its structure for modeling the multi-intersection control problems and the corresponding training algorithm. 

 \subsection{Structure of the DRI-MSC Model}
 \subsubsection{Tensor-based input}
 The traffic state information includes the number of intersections in the region, the number of cars in each lane, the distance between traffic flows and etc. In order to avoid information loss in feature extraction and avoid using a huge matrix to cover all information, we use tensors to represent the traffic state at each moment.

\begin{figure}[ht]
\centering
\includegraphics[width=0.48\linewidth]{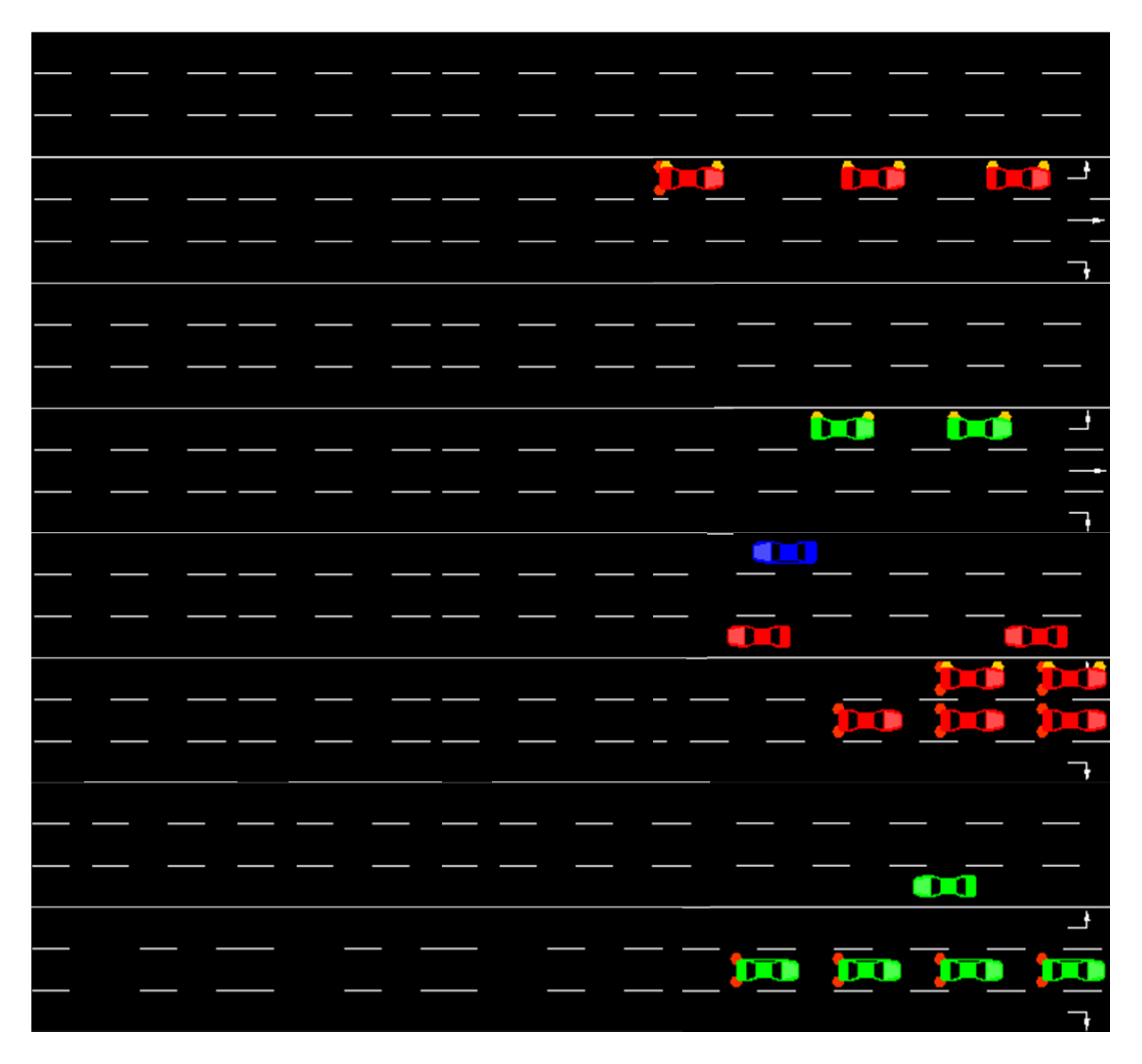}
\includegraphics[width=0.495\linewidth]{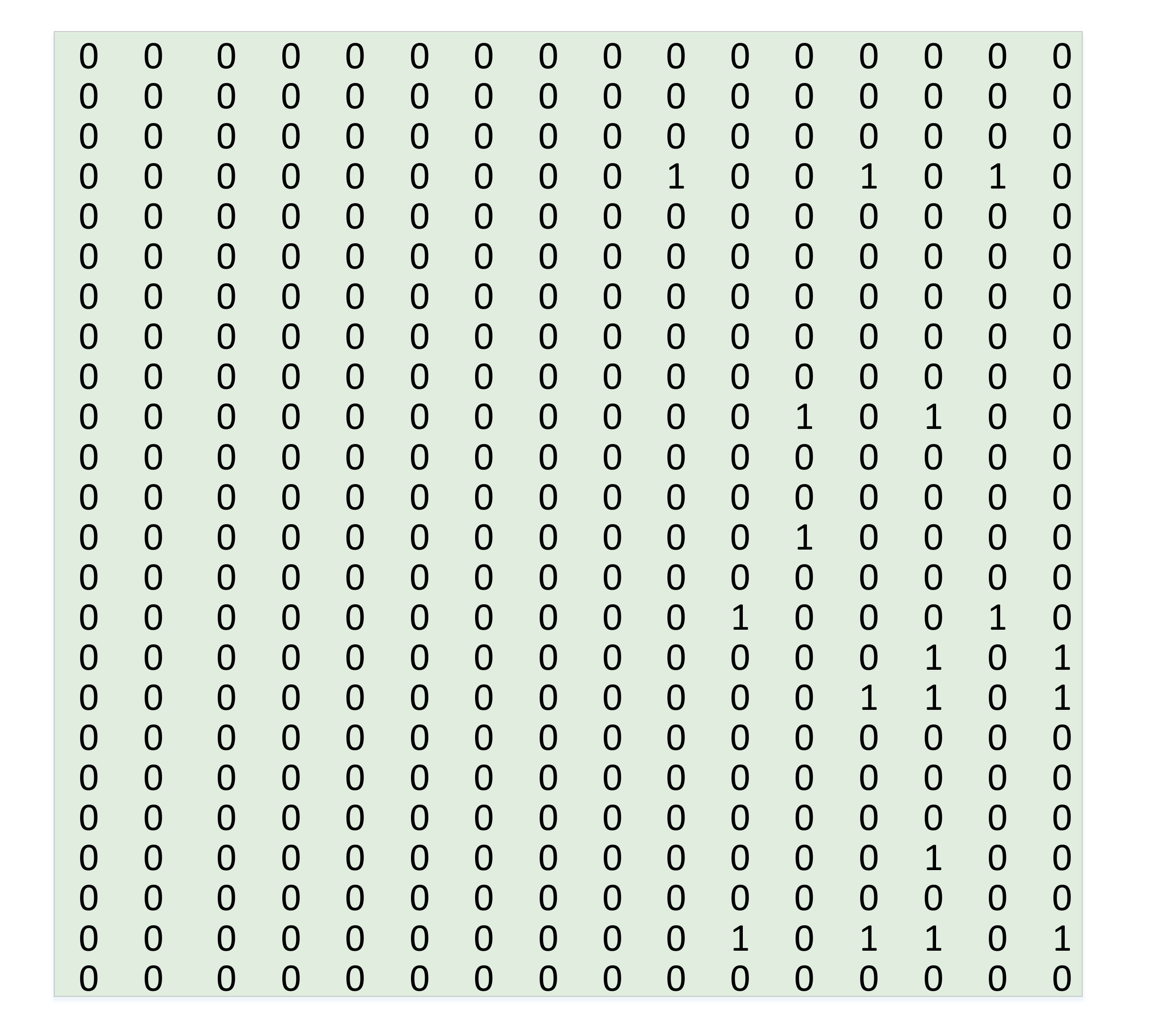}
\caption{The matrix for traffic states in a single intersection.}
\label{figmaxtrix}
\end{figure}

For a specific intersection and timestep, we get the full scene information of traffic states. In each lane, the right side is the direction close to the intersection. Thus the lanes with vehicles towards right are the entry lanes, and the lanes with vehicles facing left are the exit lanes.  Fig. \ref{figmaxtrix} shows a typical intersection with 3 lanes at each direction, and there are altogether 24 lanes from top to bottom. We then extract features by transferring  the traffic states  into a cellular matrix \cite{genders2016using}. Specifically, each lane is cut into small blocks at an interval of 5 meters, and every block is assigned  with a boolean value to present the presence and absence of a vehicle. All the values in the blocks make up of the cellular matrix, where the coordinates of each ``1'' reflect the actual position of the corresponding vehicles.

\begin{figure}[ht]
\centering
\includegraphics[width=0.48\linewidth]{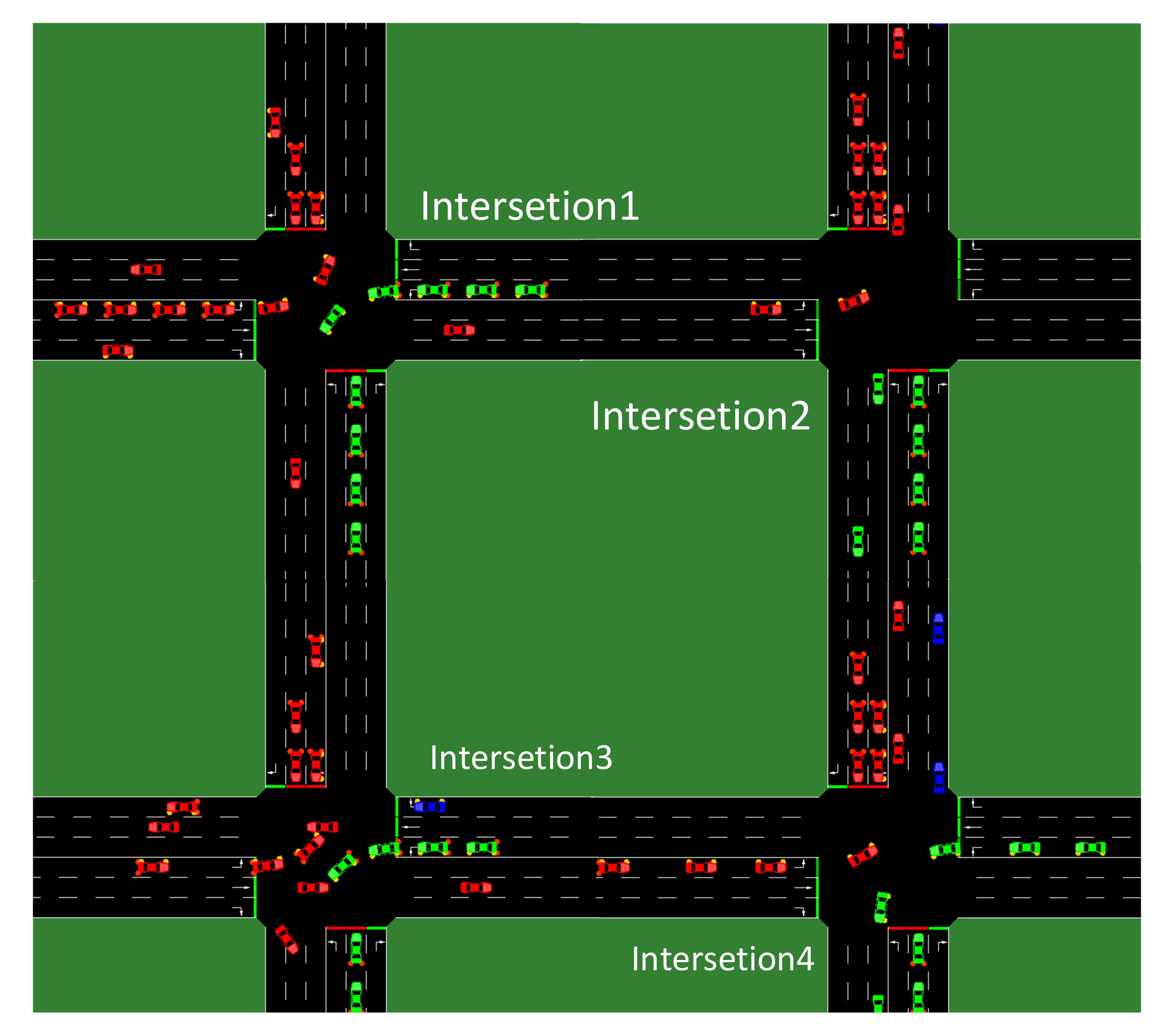}
\includegraphics[width=0.5\linewidth]{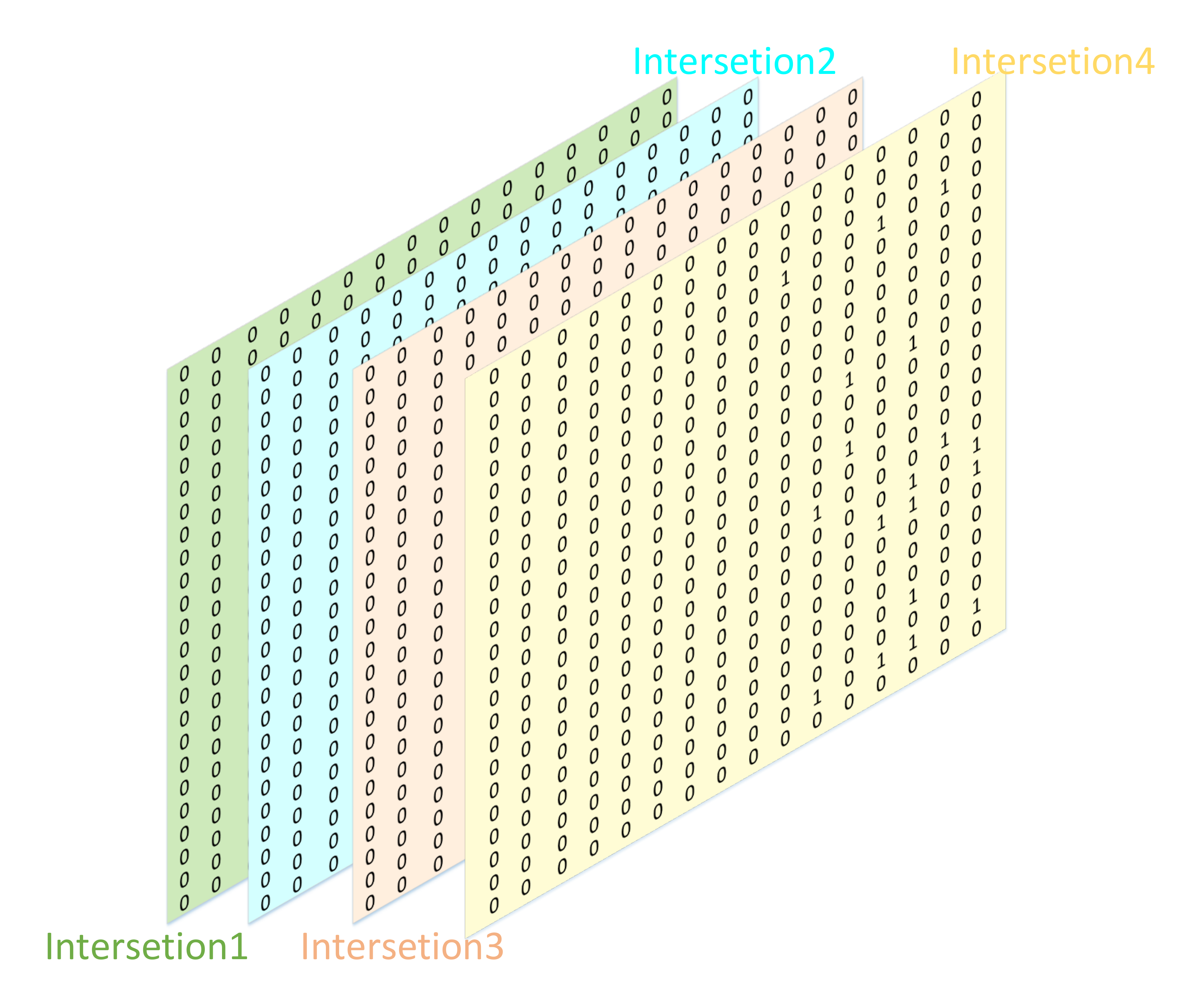}
\caption{The tensor for traffic states in 4 intersections.}
\label{figtensor}
\end{figure}

For multiple intersections, we superimpose cellular matrices of different intersections to form a tensor, which is defined as
\begin{equation}
s_t \in R^{I\times J \times K},
\end{equation}
where $K$ represents the number of blocks in each lane, $J$ is the number of lanes and $I$ denotes the number of intersections. The element $(i,j,k)$ of tensor $s_t$ is denoted by $s_{t}^{(i,j,k)}\in\left \{ 0,1 \right \}$,  in which $s_{t}^{(i,j,k)}=1$ means the presence of a vehicle in the position $(i,j,k)$. Moreover, intersections are in the position of $(i,j,K)$. For example, for the 4 intersections in Fig. \ref{figtensor}, we first get the cellular matrices for each intersection, then superimpose the matrices to form a tensor. On such basis, two convolutional layers are used to automatically extract traffic features. 

In this paper, we use a matrix to represent the phase information of the whole traffic network. which is defined as 
\begin{equation}
    h_t\in R^{I\times M},
\end{equation}
where $I$ is the number of intersections and $M$ represents the phase number of an intersection. The rows of $h_t$ represent the intersections, and the columns represent the index of phases. Element $(i,m)$  stands for the state of the $m$-th green phase of the intersection $i$, and is denoted by $h_t^{(i,m)}\in\left \{ 0,1 \right \}$. If it is the current phase, then $h_t^{(i,m)} =1$. Moreover, $M$ is equal to the phase number of the intersection with the most directions. For example, for the typical four-phase intersections involved in the experiment of this paper, $M=4$. If a T-junction exist, $``0"$ is added for unified representation. We use a dense layer to embed the phase matrix with the traffic features output by convolution layers, and then the features are transformed to outputs.

\subsubsection{output}
\begin{figure}[ht]
\centering
\includegraphics[width=1\linewidth]{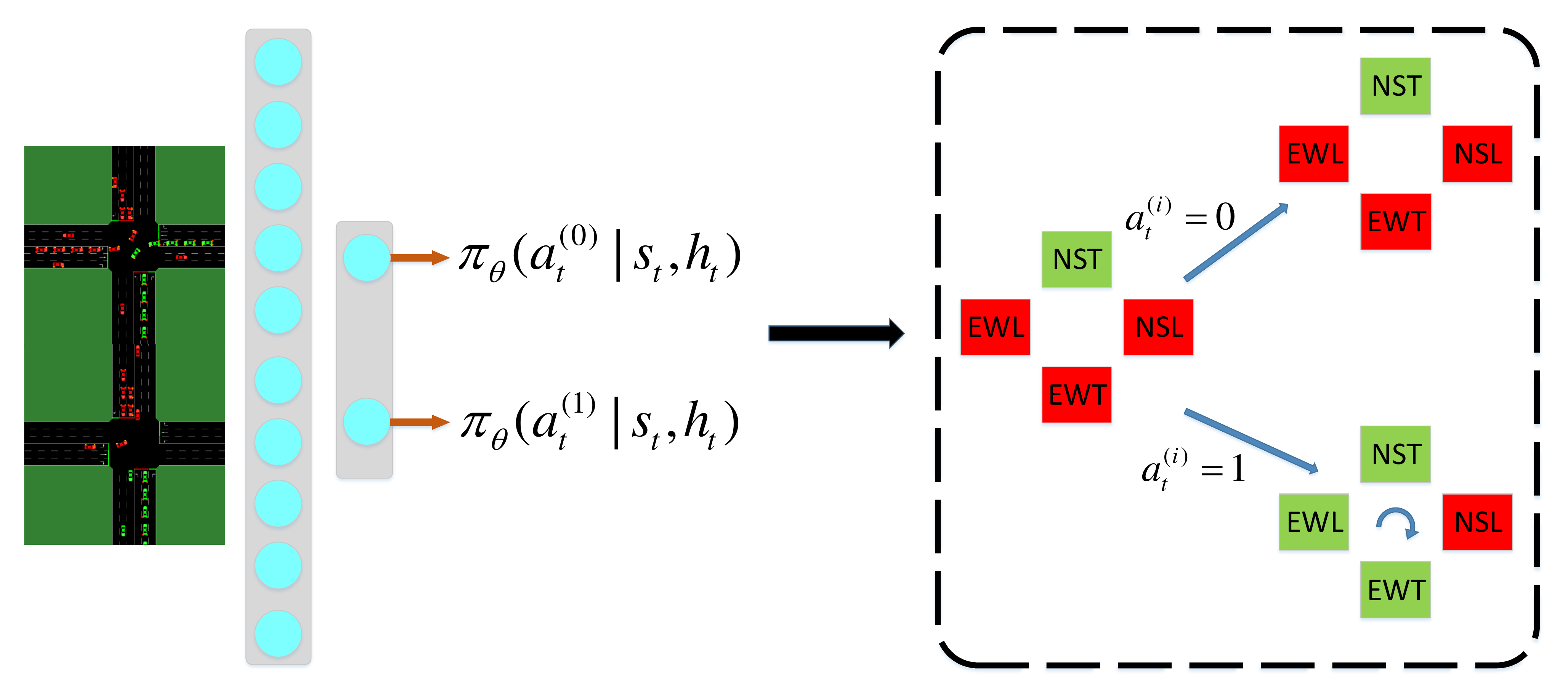}

\caption{The left shows the output of DRI-MSC model, where $\pi_\theta( a_t^{(0)}|s_t,h_t)$ and $\pi_\theta( a_t^{(1)}|s_t,h_t)$ stands for the the probability of $a_t^{(0)}$ and $a_t^{(1)}$. The right shows the phase order, where the green color stands for the green phase, the right color stands for the red phase, and the green phase changes clockwise when $ a_t^{(i)}=1$.}
\label{order_output}
\end{figure}

Although the number of phases is determined by the number of  directions, the sequence of phases based on actual rules is fixed, where we only need to decide if the green phase should be changed or not. Based on this, a binary action is sufficient to control the signal controller. Therefore, we propose  a boolean variable to represent the binary action which is whether to change the green phase or not. We name this pattern as the order-based output. Specifically, as shown in the right of Fig. \ref{order_output}, the order-based vector of multi-dimensional actions is defined as
\begin{equation}
   \pi_\theta(a_t|s_t,h_t)=f(s_t,h_t|\theta),
\end{equation}
where $f(\cdot)$ stands for the neural network, and $a_t$ is defined as the switching vector. The $i$th element of $a_t$ decide whether the green phase change for the $i$th intersection, and is denoted by $a_t^{(i)}\in\{0,1\}$, where $a_t^{(i)}=1$ means the green phase of the  $i$th intersection changing to the next direction. 

We also have $\pi_\theta(a_t|s_t,h_t)\in R^{I}$, where $I$ represents the number of intersections and $\theta$ stands for the parameters of the neural network.  The $i$th element of $\pi_\theta(a_t|s_t,h_t)$ means the probability of $a_t^{(i)}$, and is denoted by $\pi_\theta( a_t^{(i)}|s_t,h_t)\in[0,1]$,  and $\pi_\theta( a_t^{(i)}=0|s_t,h_t) + \pi_\theta( a_t^{(i)}=1|s_t,h_t)=1$. The dimension of the actions represents the policy for the corresponding intersections. Compared to the traditional phase based output,  the output of DRI-MSC model abides the practical phase changing rules, which leads to a feasibility for practical use and provides a simplified representation. 

\subsubsection{Multi-task learning structure}
\begin{figure}[ht]
\centering
\includegraphics[width=1\linewidth]{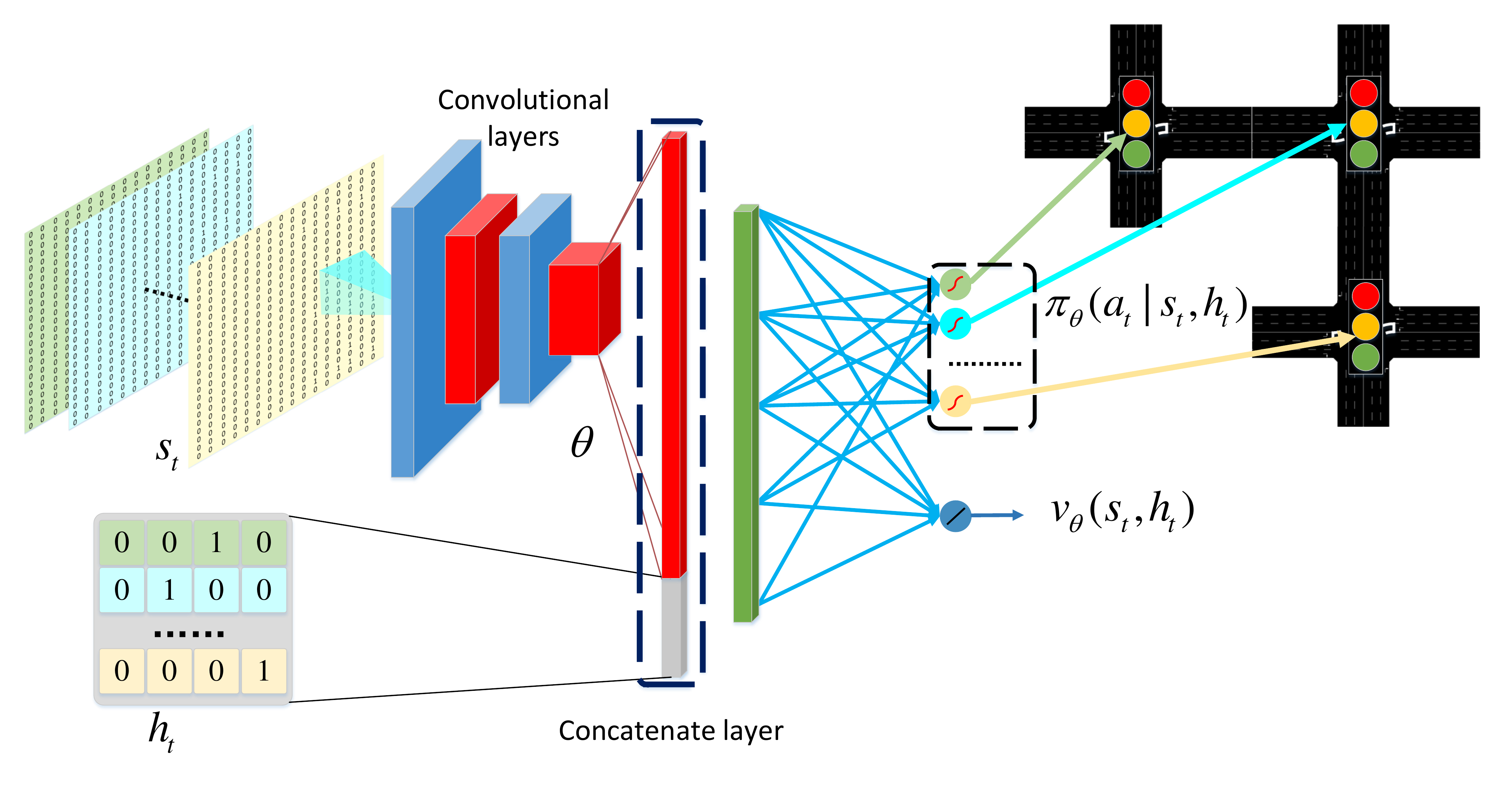}
\caption{The decision structure of the DRI-MSC model.}
\label{fig3}
\end{figure}

In robotic learning, the joints of the robots cooperate to complete complex tasks such as jumps and somersaults completely through learning \cite{2018-TOG-deepMimic}. We extend this idea to the multi-intersection traffic signal control model. Therefore, the DRI-MSC model get the collaborative policies through end-to-end multi-task learning, which is shown in Fig. \ref{fig3}. 

The main idea of this architecture is to define a unified optimization objective which regards the policy of each signal controller as a sub-task, and use a neural network with sharing parameters in multi-task architecture to output the collaborative action of each task \cite{Caruana93multitasklearning:}. Finally, a learning based method is used to train the neural network. Compared to the rule based collaborative policy, the proposed method can break through the limitation of human designing and better approximate the  optimal solution.

The  multi-task learning architecture in  the proposed DRI-MSC model  combines the policy gradient and temporal-difference errors into one, while the existing works use isolated value network and policy network and separates the total loss into two parts \cite{Silver2017Mastering,Casas2017Deep}. The DRI-MSC model can reduce the oscillation and instability while providing a more refined structure and more stable convergence. The DRI-MSC model output probabilities vector $\pi_\theta(a_t|s_t,h_t)$ and values $v_\theta(s_t, h_t)$ at the same time, where $\pi_\theta(a_t|s_t,h_t)$ is the signal control policies and $v_\theta(s_t,h_t)$ is the value of state $s_t$.

\subsection{Training method}
\subsubsection{Pre-training with imitation learning}
\begin{figure}[ht]
\centering
\includegraphics[width=0.8\linewidth]{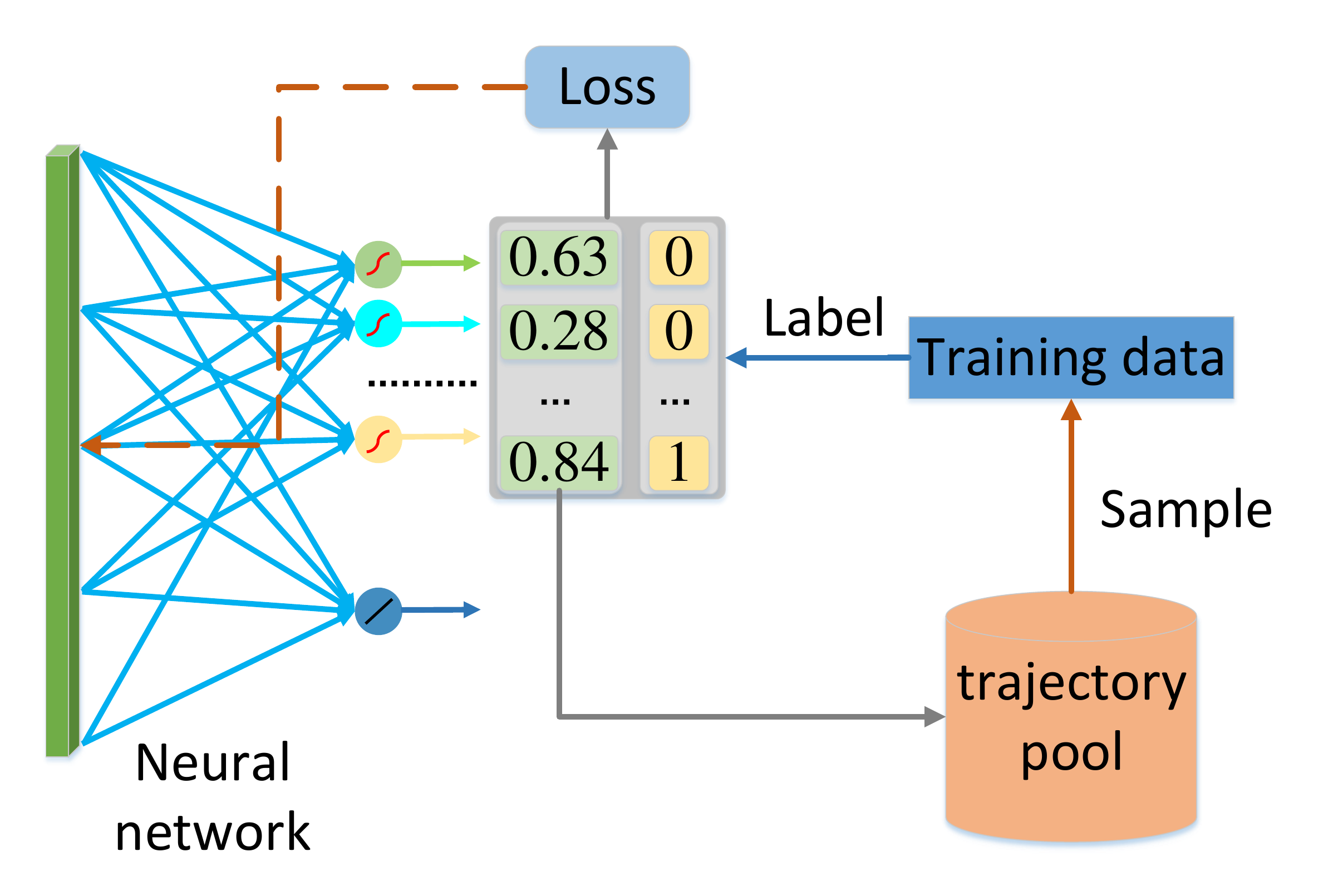}
\caption{The imitation learning process}
\label{fig4}
\end{figure}

Since the supervised learning is normally faster to converge, we pre-train the model to guarantee a warm start point for RL by imitating the policy of a rule based model. We use DAGGER algorithm  to pre-train the model, and the learning process is shown in Fig. \ref{fig4}.

When the model interacts with environment, we collect the decision trajectory data into a trajectory pool, and  randomly  sample batches of data from the trajectory pool as training samples which are labeled by a rule based multi-intersection signal control model. When the model with traffic state input and the order-based  action output has the same  the output with DRI-MSC model, it can provide a good baseline control policy. In this paper, we use the following formula to represent the model,
\begin{equation}\label{rule}
y^{(i)}=\left\{
\begin{tabular}{cc}
  0 & $\sum_{m=1}^{(M-1)}N_m^{(i)}\cdot\beta-N_0^{(i)}\leq0$\\
  1 & $\sum_{m=1}^{(M-1)}N_m^{(i)}\cdot\beta-N_0^{(i)}>0$
\end{tabular}
\right.
\end{equation}
where $y^{(i)}$ represents the label of the $i$th dimension, which is also the rule-based policy corresponding to the traffic state of the intersection $i$. Besides, $N_0^{(i)}$ indicates the number of low-speed vehicles in lane $i$ with green phase, and   $\sum_{m=1}^{(M-1)}N_0^{(i)}$  indicates the number of low-speed vehicles in the lanes with red phase, and $M$ is the phase number. We use the number of vehicles at low speed to roughly reflect the size of traffic flow and congestion. The core of (\ref{rule}) is that if the direction with green phase is much more crowded than other directions, then we keep the existing phase, otherwise it switches to the next phase. The advantage of such settings lies on the fact that the policy model is very small and  is easy to calculate and achieve a certain result. In addition, in this paper, the vehicles with a speed lower than 30km/h are defined as a low-speed vehicle, and we set $\beta=0.13$.

The loss of the imitation learning model is supposed to reflect all the intersections, thus we define the global loss as the sum of errors in each dimension, which is formulated as
\begin{equation}\label{global_loss}
    E=\sum_{t=0}^{T-1}\sum_{i=0}^{I-1}e_t^{(i)}+c\left \| \theta \right \|^2,
\end{equation}
where $e_t^{(i)}=y_t^{(i)}\ln(\pi_\theta( a_t^{(i)}|s_t,h_t))+(1-y_t^{(i)})\ln(1-\pi_\theta( a_t^{(i)}|s_t,h_t))$ and $y_t^{(i)}$ represents the label of the first dimension. $T$ means  the  terminal  time-step defined in (\ref{value_equ}). $E$ is designed to measure the similarity between the model outputs and the sample labels. In addition, we added $l_2$ norm penalty  to avoid over-fitting. Then we employ the stochastic gradient descent algorithm for the  optimization.
\subsubsection{Fine-tuning with RL}
After the pre-training  to accelerate the convergence, RL method is then used to fine tune the model.
\begin{figure}[t]
\centering\includegraphics[width=0.9\linewidth]{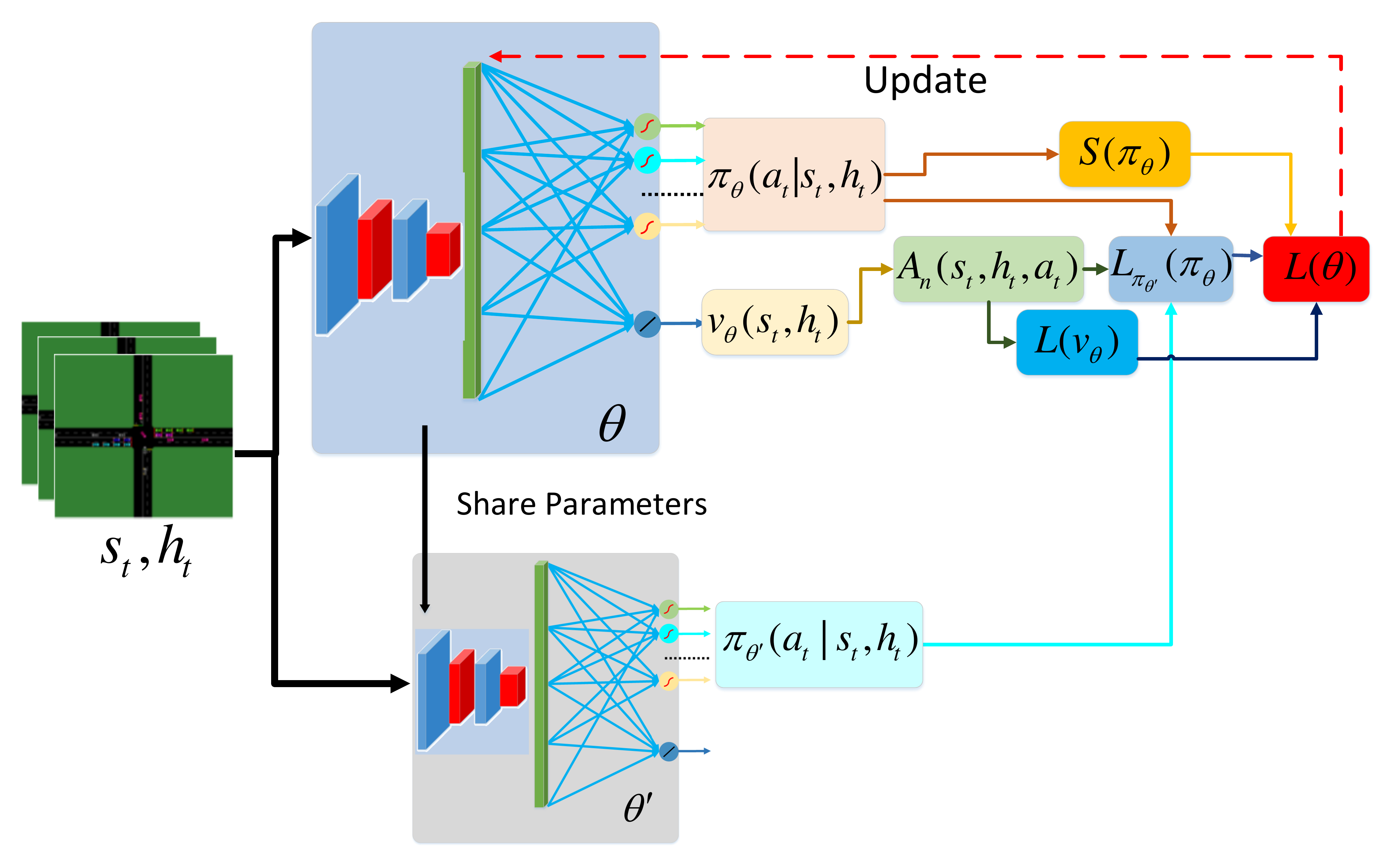}
\caption{The reinforcement learning process}
\label{rl}
\end{figure}

For a RL based model, the reward function gives a quantitative description of the optimization objective and a evaluation criterion for policies and actions. In a traffic network, since adjacent intersections influence each other, individual optimization on each intersection can lead to the prisoner's dilemma for the whole network \cite{rapoport1965prisoner}. Previous works either use the change in cumulative vehicle delay, or use a simplified queue length  as reward \cite{genders2016using,Zeng2018AdaptiveTS}. However, delay is a lagging indicator and convergence rate can be negatively impacted. The coming flow and slow-moving vehicles can be ignored when employing simplified queue length. Due to the complexity of traffic states and the mutual influencing mechanism among intersections, we directly describe the total spatial objective by setting the reward function as the change of  numbers of low-speed vehicles in all lanes of the entire traffic network, 
\begin{equation}
    R_t=\sum_i(N_t^{(i)}-N_{t+1}^{(i)}),
\end{equation}
where $N_t^{(i)}$ represents the number of low-speed vehicles in lane $i$ at time $t$. In other words, we use the changes in low-speed vehicles to reflect the effect of the policies. When traffic conditions improve, the number of low-speed vehicles decreases and the model receives a positive reward and vice verse. With this reward function, real-time traffic density information and the objective can be strongly related.

For actor critic, $v_{\theta}(s_t,h_t)$ is used to estimate the real value $v_\pi(s_t,h_t)$ defined in (\ref{value_equ}), and $A_n(s_t,h_t,a_t)$ is defined as 
\begin{equation}\label{multistep_esti}
A_n(s_t,h_t,a_t)=\sum_{k=1}^n(R_{t+k}+\gamma^kv_\theta(s_{t+n},h_{t+n}))-v_\theta(s_t, h_t),
\end{equation}
which is used to estimate $A^\pi(s_t,h_t,a_t)$ in (\ref{policy_formula}) based on bellman equation \cite{bellman1954}.

In this paper,  different time steps $A_1(s_t,h_t,a_t),A_2(s_t,h_t,a_t),\cdots A_N(s_t,h_t,a_t)$ are used to make a proper trade off between short-term and long-term rewards  and $N$ stands for the max step \cite{Schulman2015HighDimensionalCC}. Moreover, $A_n(s_t,h_t,a_t)$ can be regarded as temporal-difference error, which can be used to calculate the error of the value network \cite{Tesauro:1995:TDL:203330.203343}. Then, the loss function of the estimated value $v_\theta(s_t,h_t)$ is shown as follows
 \begin{equation}
L(v_\theta(s_t,h_t))=\frac{1}{N}\sum_{n=1}^{N}(A_n(s_t,h_t,a_t)).
\end{equation}

For the multi-dimensional action model in this paper, the convergence remains a problem due to the mismatch between parameter space and action space and the high coupling between policies and environment \cite{Kakade:2002:AOA:645531.656005}. A solution to this problem is to limit the update magnitude of a policy by adding constraints \cite{pmlr-v37-schulman15,schulman2017proximal,Achiam:2017:CPO:3305381.3305384}. In this paper, we use PPO method for its advantage in computation consumption. The main idea of PPO is to strict the magnitudes of the parameters updates by truncating the policies. The target function corresponding to this method can be expressed as follows
\begin{equation}\label{equ5}
\begin{array}{ll}
    L_{\pi_{\theta'}}(\pi_{\theta})=&E_{\tau\sim\pi_{\theta'}}[\min(A_n(s_t,h_t,a_t),\\
   								& \text{clip}(r_t^{\pi_{\theta'}}(\pi_{\theta}), 1-\varepsilon ,1+\varepsilon )A_n(s_t,h_t,a_t))],
\end{array}
\end{equation}
with
\begin{equation}
    r_t^{\pi_{\theta'}}(\pi_{\theta})=\frac{\pi_{\theta}(a_t\mid s_t,h_t)}{\pi_{\theta'}(a_t\mid s_t,h_t)},
\end{equation}
 where $\pi_\theta$ and $\pi_{\theta'}$ stands for the policy based on $\theta$ and $\theta'$, and $ r_t^{\pi_{\theta'}}(\pi_{\theta})$ is the ratio of the new policy to the old policy. Moreover, $\varepsilon$ is the threshold of policy change, and $ \text{clip}( r_t^{\pi_{\theta'}}(\pi_{\theta}),1-\varepsilon ,1+\varepsilon )$ means clip the probability ratio $r_t^{\pi_\theta}(\pi_{\theta'})$ to a range of $[1-\varepsilon ,1+\varepsilon ]$.

Fig.\ref{rl} gives the illustration of the RL process used in this paper. The total objective function $L(\theta)$ is as
 \begin{equation}\label{rl_loss}
    L(\theta)=L_{\pi_\theta}(\pi_{\theta'}))-\alpha_1L(v_\theta)+\alpha_2S(\pi_\theta),
 \end{equation}
where the policy function $ L_{\pi_\theta}(\pi_{\theta'})$ can be computed by (\ref{equ5}) and value function $L(v_\theta)$ can be computed by (\ref{multistep_esti}). $S(\pi_\theta)$ is the entropy of the policy $\pi_\theta$, which is a soft RL based regularization method \cite{Ziebart:2010:MPA:2049078}. The main idea of soft RL is to succeed at the task while acting as randomly as possible. 
Experiments show that this method can improve the stability and scalability of RL model\cite{haarnoja2018soft,pmlr-v48-mniha16}.

Then the new policy $\pi_\theta$ of this iteration (or the old policy $\pi_{\theta'}$ of the next iteration) can be obtained by maximizing $L(\theta)$:
\begin{equation}
    \pi_{\theta'}\leftarrow \underset{\pi_\theta}{{\rm argmax}}(L(\theta))
\end{equation}
Fig. \ref{rl}. shows the complete parameter update process, where the policy is improved iteratively. 

\subsubsection{Co-training}
The details of the co-training process are shown in Algorithm \ref{alg:A}. In the first stage of training, we use the accuracy $Acc$ defined as 
 \begin{equation}\label{Acc}
    Acc=\frac{\sum_{t=0}^{T-1}\sum_{i=0}^{I-1}\left |  a_t^{(i)}-y_t^{(i)}\right |}{T+I}.
 \end{equation}
to evaluate the model output. Since imitation learning reduces the searching space simultaneously, exorbitant $Acc$ can lead to a negative impact on RL performance. Therefore, we introduce a early stop mechanism, which sets a threshold $\xi$ to control the imitation accuracy. The imitation learning process ends when  $Acc$ reaches  $\xi$, and then the reinforcement learning process start.

\begin{algorithm}[ht]
\caption{DRI-MSC Co-training Algorithm}
\label{alg:A}
\begin{algorithmic}
\STATE {Randomly initialize new policy network $\pi_{\theta}$, old policy network $\pi_{\theta'}$ with weights $\theta$ and $\theta'$, the traffic state $s_1$ and the trajectory pool $D$.
}
\REPEAT
\FOR{$t=1\rightarrow T$}
\STATE Choose action $a_t$ based on policy $\pi_{\theta}$;
\STATE Execute $a_t$ and observe the next state $s_{t+1}$,$h_{t+1}$;
\STATE Store $s_{t+1},h_{t+1}$ in $D$;
\STATE Sample a random minibatch of trajectory data $s_{r1},h_{r1},\ s_{r2},\ h_{r2},\cdots,\ s_{rN_b},\ h_{rN_b}$;
\STATE Compute lables $y_{r1},\ y_{r2},\cdots, \ y_{rN_b}$;
\STATE Train the new policy network with respect to $\theta$ for $m$ iterations.
\ENDFOR
\UNTIL{$Acc>\xi$}
\FOR{$n=1\rightarrow N_{rl}$}
\FOR{$t=1\rightarrow T$}
\STATE Choose action $a_t$ based on policy $\pi_\theta$;
\STATE Execute $a_t$ and observe the next state $s_{t+1}, h_{t+1}$;
\STATE Store transition $(s_t,h_t,a_t,r_t,s_{t+1},h_{t+1})$  in buffer.
\FOR{every $N$ steps}
\STATE Update the weights of old policy network $\theta'$ with  $\theta$
\STATE Compute advantage function $A_1(s_t,h_t,a_t), \ A_2(s_t,h_t,a_t),\cdots, \ A_N(s_t,h_t,a_t)$
\STATE Update the new policy network by maximizing $L(\theta)$
\STATE Clear buffer.
\ENDFOR
\ENDFOR
\ENDFOR
\end{algorithmic}
\end{algorithm}

\section{Numerical experiment}\label{sect3}
In this section, a series of numerical experiments are conducted to evaluate the proposed DRI-MSC model from different aspects. Under different scenes and flows, comparisons with related state-of-art methods are presented to illustrate the efficacy and advantages of the proposed DRI-MSC model in tackling large scale control problem for multi-intersections traffic signal. Then detailed experiments and analysis towards the specific policies in DRI-MSC model are then discussed to give further insights.
\subsection{The basic settings and assumptions}
\begin{figure}[ht]
\begin{subfigure}{0.5\textwidth}
\centering
\includegraphics[width=0.6\linewidth]{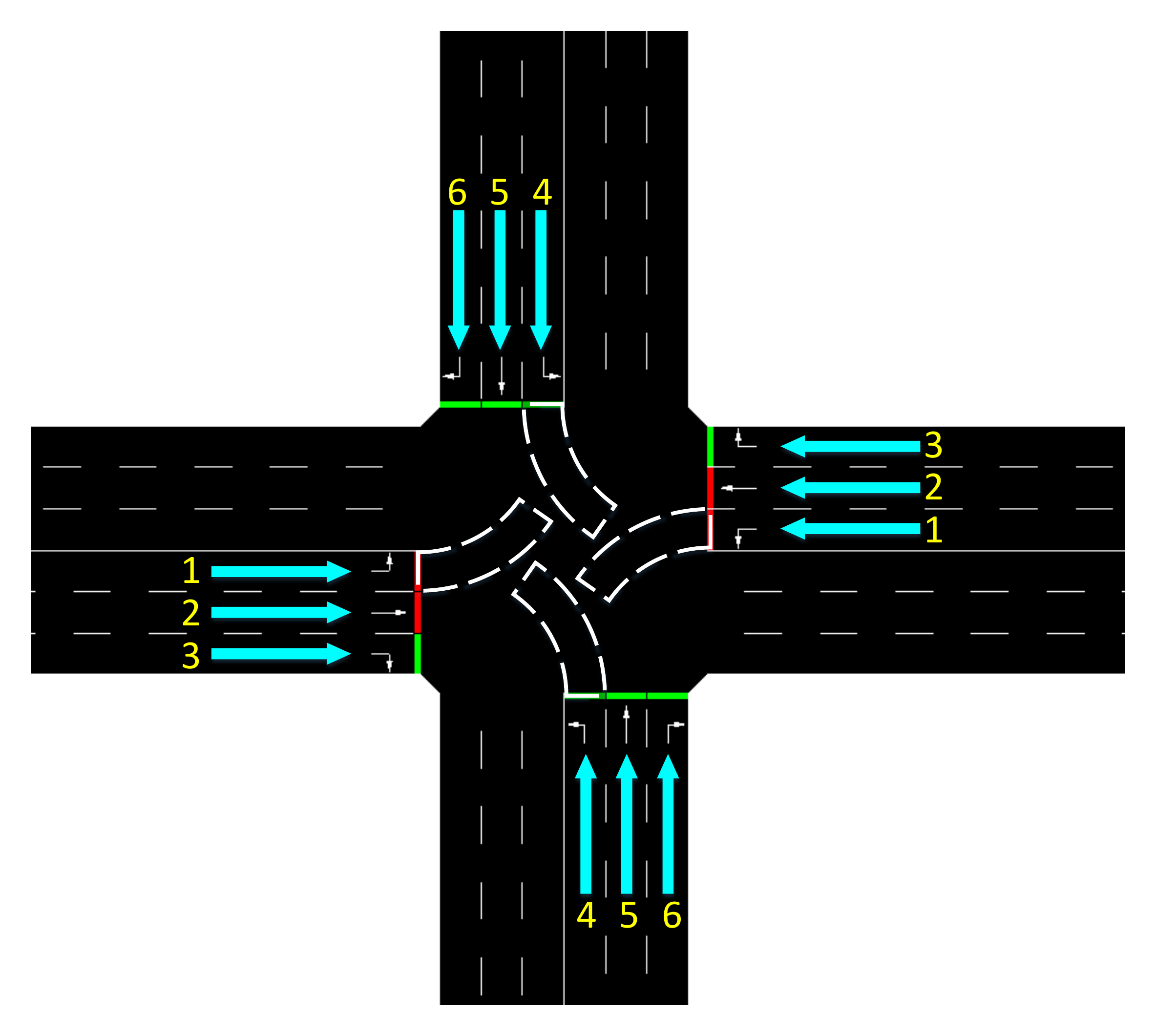}
\caption{Isolated intersection.}
\label{fig_6a}
\end{subfigure}
\begin{subfigure}{0.5\textwidth}
\centering
\includegraphics[width=0.6\linewidth]{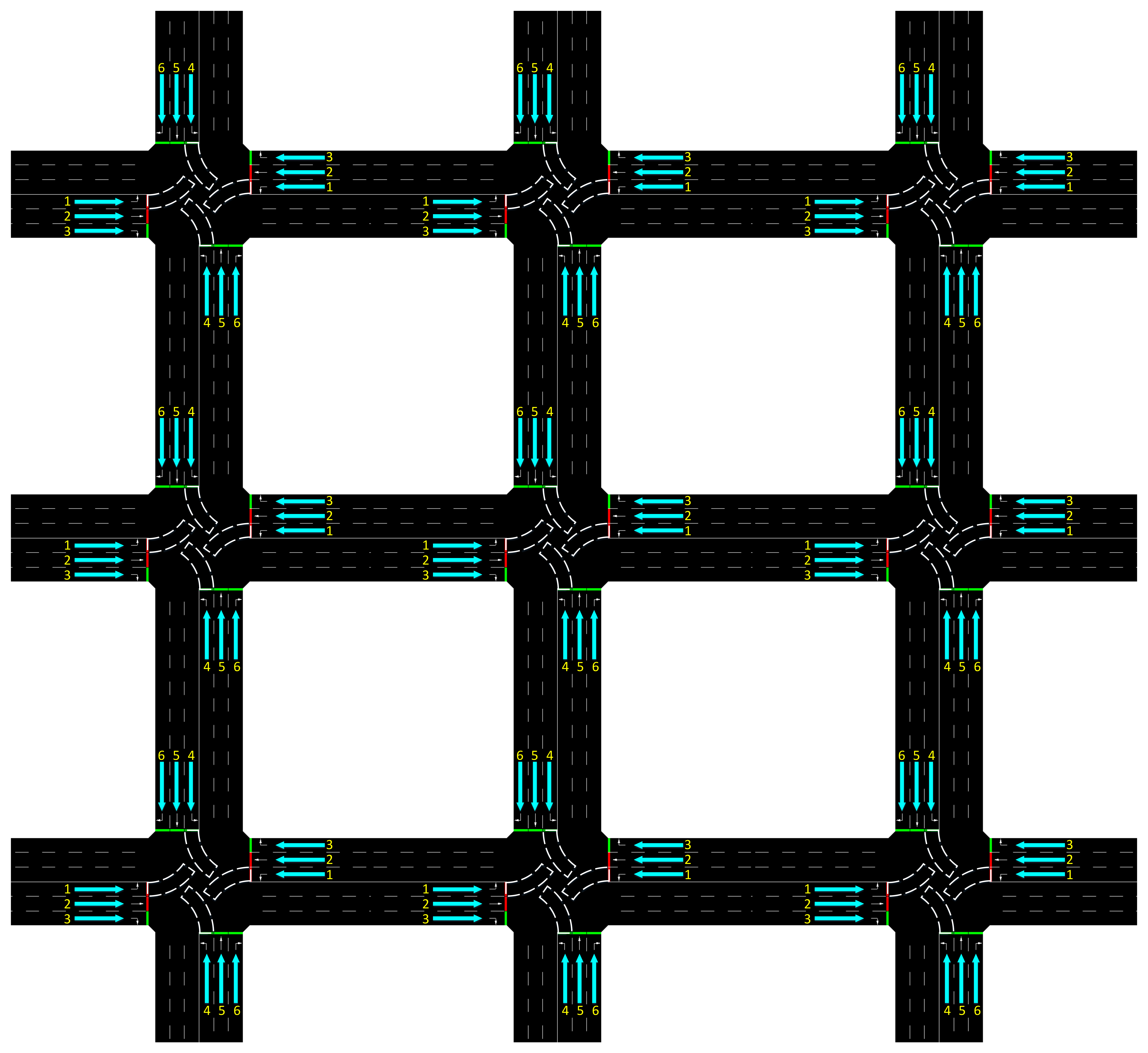}
\caption{Traffic network with 9 intersections.}
\label{fig_6e}
\end{subfigure}
\caption{Illustration of experimental scenes.}
\label{fig6}
\end{figure}

The simulation of urban mobility (SUMO) platform is taken as the experimental environment, in which we standardize the experimental environment and objects. Specifically, all intersections are typical 4-way, the number of lanes is set as 3 in each direction, the width of each lane as 3.5 meters (m), the maximum speed of the lane as 50 kilometer per hour (km/h), and the length of road as 500m in all intersections. For the intersections, as shown in Fig. \ref{fig_6a}, the rightmost lanes are right-turn lanes whose phases are always green, and the middle lanes are through lanes.

For the phase setting, the phase order is first north-south straight, north-south left, then east-west straight, and last east-west left. In order to ensure that the model is similar to the real world and the traffic safety is guaranteed. Amber phase with the duration of one timestep is inserted between two different phases, and the model outputs a policy at each timestep.

We assume that the information of each car in all lanes can be obtained, all the traffic flows are composed of cars which abide  traffic rules,  and the traffic flow  is uniformly distributed in the system. In addition, pedestrians are not considered, and a left turn waiting area is set. For the experiment setting, we set the duration of each simulation episode as 4000 seconds (s). All the experiments are conducted on Intel(R) Core(TM) i7-6900k CPU @3.20GHz with NVIDIA GTX TITAN Xp Pascal.

Fig. \ref{fig6} shows the experimental scenes  in the paper. Table \ref{flows} describes the flow data in the single-intersection experiment, where  ``S'' represents one simulation episode, and ``n/S'' is short for the number of vehicles in every simulation episodes. The direction of flows in Fig. \ref{fig_6a} is indicated by arrows indexed with numbers. For example, the arrow ``1'' indicates the direction of route 1, and represents the flow in the east-west direction. Without specific notation, the traffic data at multiple intersections are all low flows.

\begin{table*}[ht]
\caption{Illustration of experimental flows.}\label{flows}
\begin{tabular}{ m{3cm} m{1.4cm} m{1.4cm}  m{1.4cm} m{1.4cm} m{1.4cm}  m{1.4cm} m{1.4cm} m{1.4cm} }
\Xhline{1.2pt}
 Flow & 1 & 2  &3 & 4 & 5  &6 &Time & Total \\
         					    & (n/S) & (n/S) & (n/S) & (n/S) & (n/S) & (n/S) & (S) &\\
\hline
Low & 600 & 600 &600 &600 &600 &600 &1 &3600
\\
Middle &800 &800 &800 &800 &800 &800 &1 &4800
\\
High &1000 &1000 &1000 &1000 &1000 &1000 &1 &6000
\\
					& 600 & 600 &600 &600 &600 &600 &1/3 &1200\\
Mutable     &900 &900 &900 &900 &900 &900 &1/3 &1600\\
					&720 &720 &720 &720 &720 &720 &1/3 &1440\\
 						& 600 & 900 & 600 &900 &600 &600 &1/3 &1400\\
Unbalanced	 &900 &600 &600 &600 &900 &600 &1/3 &1400\\
						&900 &900 &600 &600 &900 &600 &1/3 &1400\\
\Xhline{1.2pt}
\end{tabular}
\end{table*}

All the hyper-parameters of the neural network of DRI-MSC model are show in Table \ref{table1}. The layers and units are minimized with the imitation accuracy guaranteed. The DRI-MSC model contains two convolutional layers to extract features of traffic states from the input tensors and two pooling layers to reduce dimensionality \cite{726791}. Besides, we combine them with phases information, and map it to the output of the model with a fully connection layer.
\begin{table}[ht]
\caption{Parameters of the neural network architecture.} \label{table1}
\begin{tabular}{ c c c c }
\Xhline{1.2pt}
Layers & Kernel Size & Number &Activation\\
\hline
Convolution layer 1 & $5\times5$& 32 &Relu\\
Max pooling layer 1 & $1\times2$&    &Linear\\
Convolution layer 2 & $3\times3$& 64 &Relu\\
Max pooling layer 2 & $2\times2$&    &Linear\\
Fully-Connected layer & $3\times3$& 500 &Relu\\
Output Layer & & $Num+1$ &Sigmoid/Linear\\
\Xhline{1.2pt}
\end{tabular}
\end{table}

\subsection{Training of DRI-MSC Model}
For a network composed of $Num$ intersections, the number of neurons in the output layer is $Num+1$, in which the sigmoid activation function maps the probability to control the traffic light at each intersection and the  value of the traffic state $v(s_t)$ is also outputted additionally to compute policy gradient. 

For the imitation learning process, we set the coefficient $c$ of (\ref{global_loss}) as $10^{-4}$. In Algorithm \ref{alg:A}, after one simulation episode (4000s), $m=500$ training iterations are conducted with batch size $N_b=100$, and the threshold $\xi=0.9$ when facing the isolated intersection environment. In the multi-intersection environment, we set $\xi=0.7$ to encourage more exploration. For the RL process, we set $\gamma$ in (\ref{value_equ}) as $0.6$. In equation (\ref{rl_loss}), $\alpha_2$ is set to $0.1$ to control the convergence speed and balance exploration with exploitation. The coefficient $\alpha_1$ is then employed to balance the priority of policy objective and value objective. We found in  the experiments that larger and smaller $\alpha_1$ both have a negative influence on convergence, and $\alpha_1=1$ is set as a satisfactory trade off.

For imitation learning based model, we use imitation loss and accuracy to the rule based model as evaluation indexes. For RL based model, we use entropy as value loss to evaluate the convergence. For a traffic control model, indexes such as queue length, average waiting time, average speed and energy consumption per vehicle can reflect the efficiency, and we adopt average phase duration as another evaluation index.

\subsection{Experimental results and analysis}
\begin{figure}[t]
\begin{subfigure}{0.5\textwidth}
\centering
\includegraphics[width=0.9\linewidth]{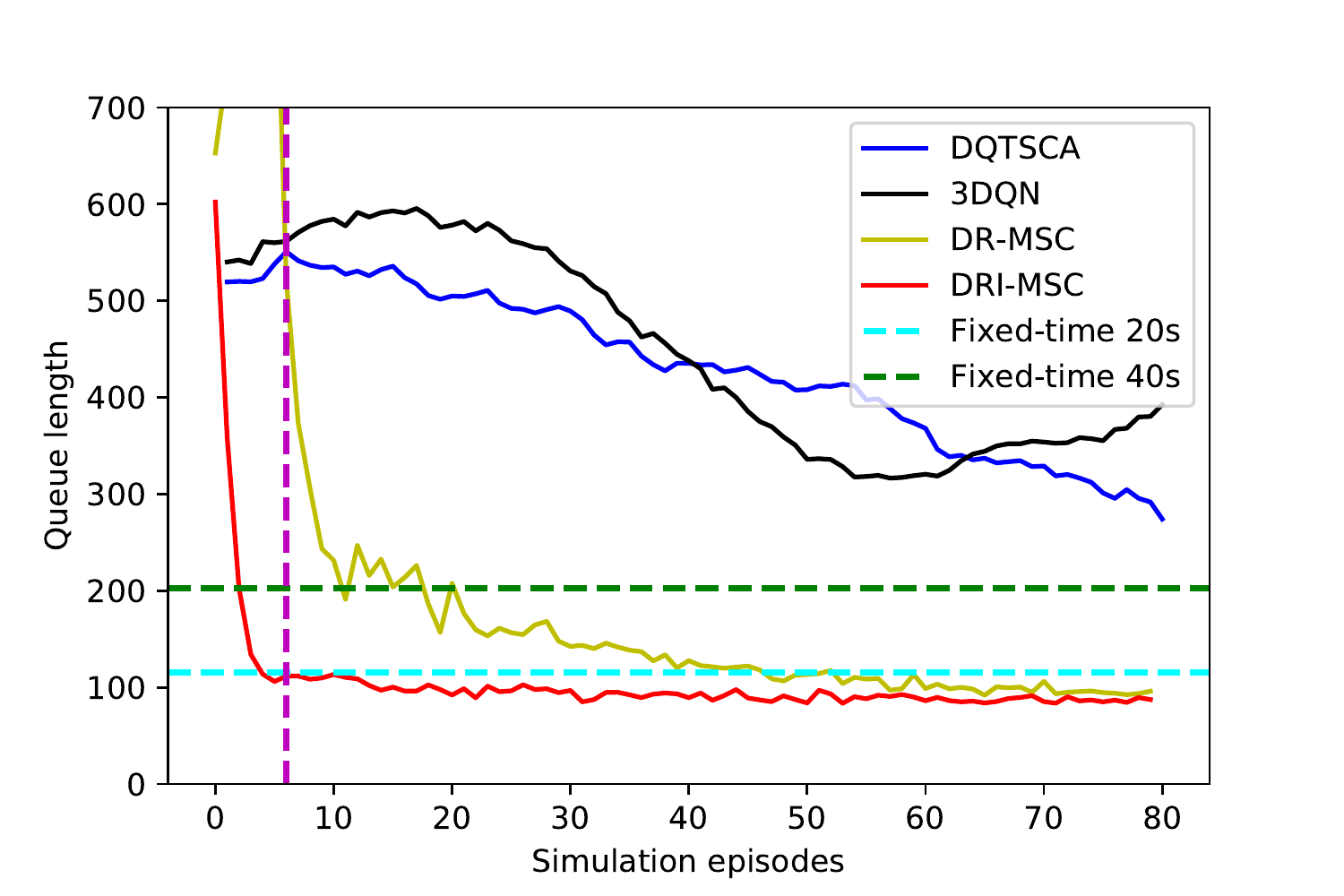}
\caption{The comparison of models for isolated intersections.}
\label{diff_model}
\end{subfigure}
\begin{subfigure}{0.5\textwidth}
\centering
\includegraphics[width=0.9\linewidth]{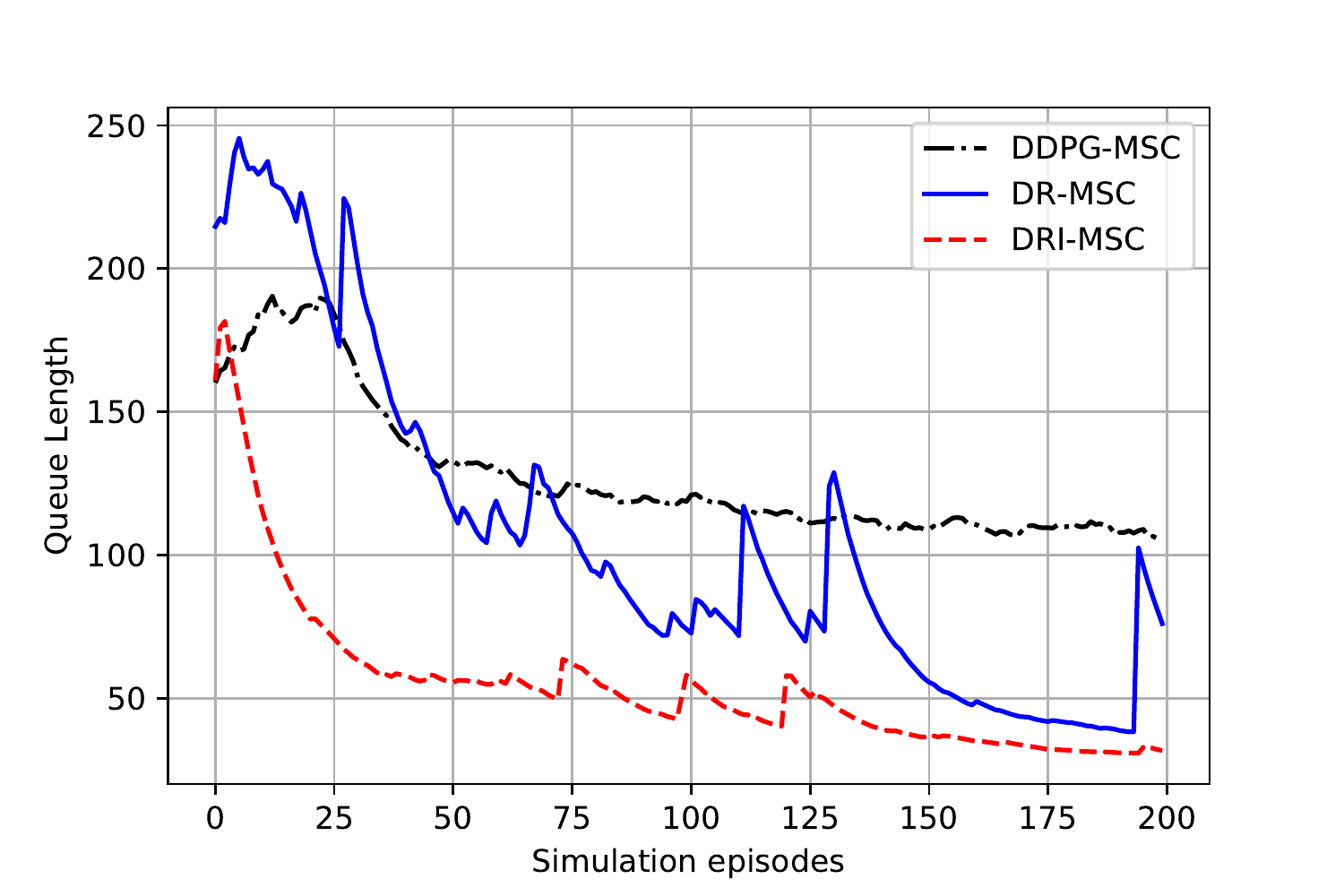}
\caption{The comparison of models for 4-intersection.}
\label{com_ddpg}
\end{subfigure}
\caption{The comparison with other stare-of-art methods.}
\end{figure}

We first compare the convergence of DRI-MSC model with other methods, where the deep Q-network traffic signal control agent (DQTSCA) model, double dueling deep Q-network (3DQN) model and two fixed-time control policies are introduced for the isolated intersection  \cite{genders2016using,liang2018deep}. For multi-intersection, we compare the proposed DRI-MSC model with the deep deterministic policy gradient (DDPG) model, which is a benchmark deep RL model applicable for multi-dimensional control. For the DDPG based model, we set the capacity of the experience replay to 10000, and softly assign the parameters with a rate of 0.01. All the other parameters are the same as DRI-MSC model. Since DQTSCA model and 3DQN are only based on deep RL, we also compare  the DRI-MSC model without imitation learning (DR-MSC) to test the effect of the RL and imitation learning methods in our proposed model.  Fig. \ref{diff_model} gives the comparison results.

Fig. \ref{diff_model} shows that the queue length of DQTSCA and 3DQN models need hundreds of simulation episodes to converge to the level of fix-time policies, while the DRI-MSC model and the DR-MSC model only needs several episodes to converge by contrast. For the 4-intersection experiment in Fig \ref{com_ddpg}, it shows that the DRI-MSC model converges faster than the DDPG-MSC model. The DR-MSC model has a similar convergence speed with that of the DDPG-MSC model at the initial epochs, and then DR-MSC model also converges faster with epochs going on. Except the convergence speed, we can also see that the proposed DRI-MSC model can reach a better result, i.e., a shorter queue length. Fig. \ref{diff_model} and Fig \ref{com_ddpg} verify that the proposed DRI-MSC model outperforms the given methods  both in convergence speed and final performance.  Besides, Fig \ref{com_ddpg}  also illustrates that imitation learning method accelerates the converging process in the proposed DRI-MSC model.

\begin{figure}[ht]
\begin{subfigure}{0.5\textwidth}
\centering
\includegraphics[width=1\linewidth]{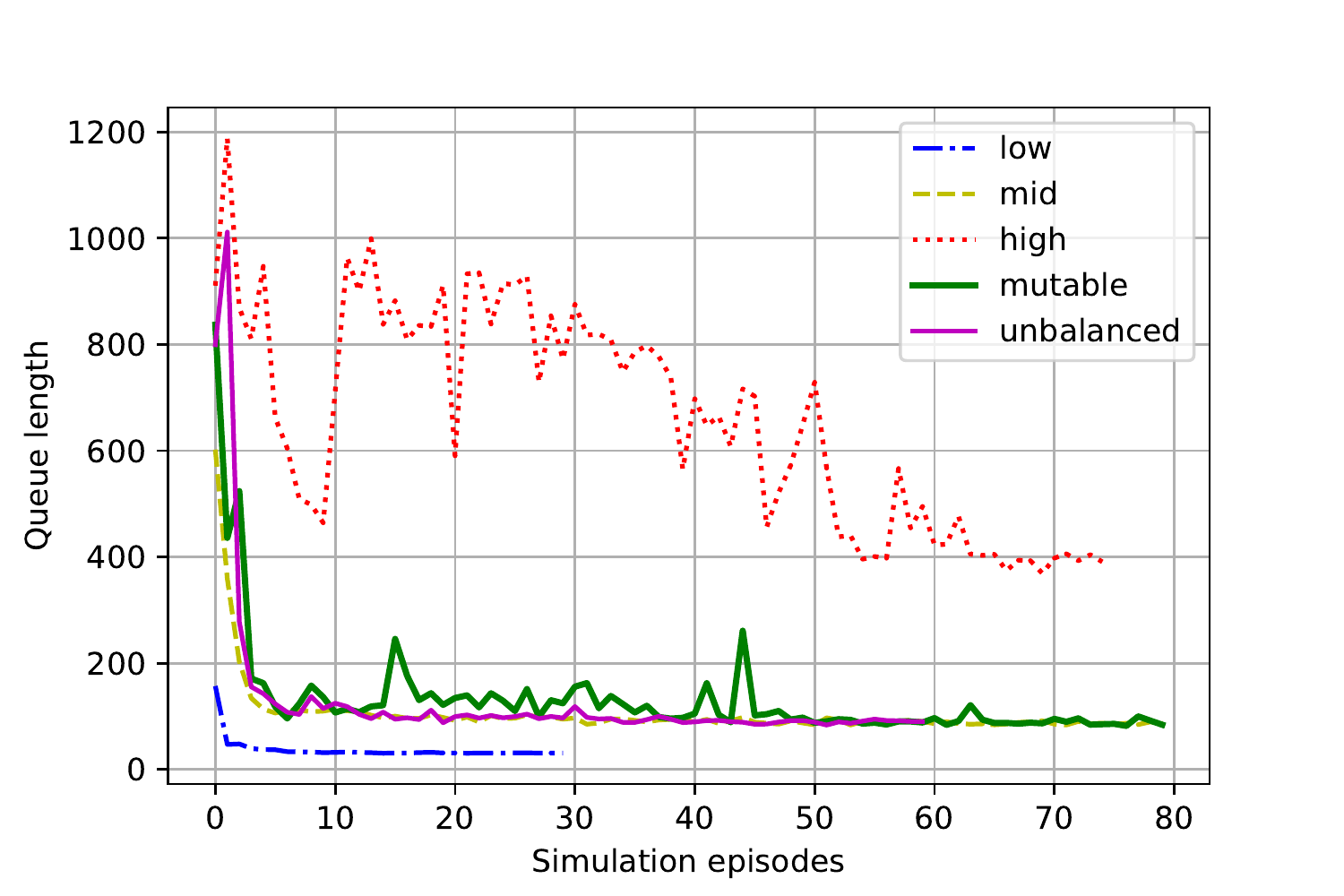}
\caption{Queue length of different flow.}
\label{fig_9a}
\end{subfigure}
\begin{subfigure}{0.5\textwidth}
\centering
\includegraphics[width=1\linewidth]{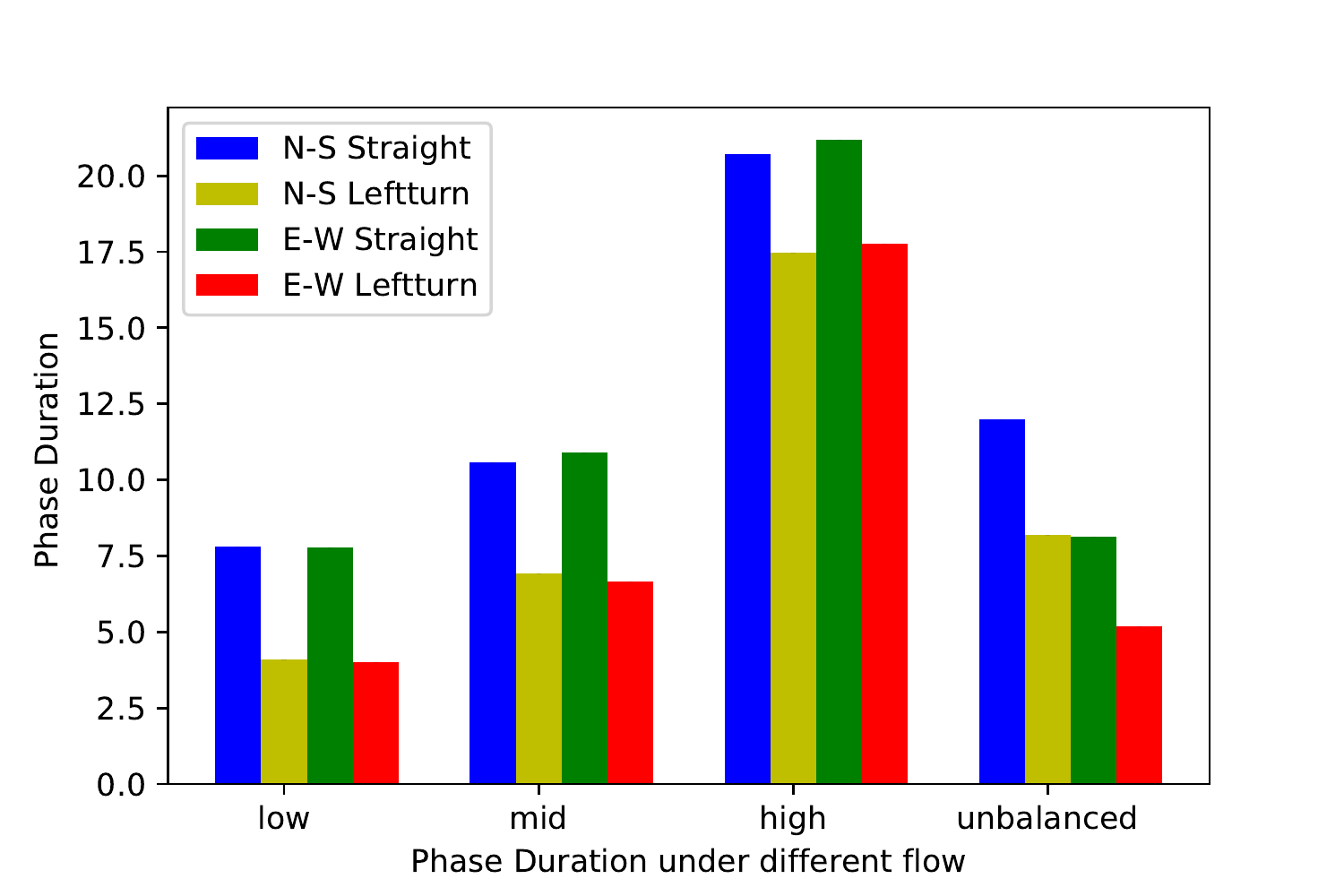}
\caption{Queue length of different intersection numbers}
\label{fig_9b}
\end{subfigure}
\caption{The phase duration under different flows, in which ``N", ``S", ``E"  and ``W" represent the north, south, east and west respectively.}
\label{fig_9}
\end{figure}

After verifying the advantages of the DRI-MSC model, we then evaluate it overall performance under different conditions and from different aspects, together with corresponding analysis. We first evaluate the DRI-MSC model in Fig \ref{fig_9} with different flows given shown in Table \ref{flows}. 

Fig. \ref{fig_9a} illustrates that the DRI-MSC model converges in all the flows given. We can see that the model converges slower when the traffic become heavier, which is reasonable. Under the low flow condition, the model can converge to a good point within two episodes of simulation. Except high flow, the model converges to a good point within less than 7 hours of simulation time. Therefore, the DRI-MSC model based policy is more acceptable and applicable when compared with other RL models, since the ordinary RL models require up to months of  training cost. Moreover, mutable flow and asymmetric flow do not significantly increase the training time under the condition of a certain traffic intensity, which shows good adaptability of the proposed DRI-MSC model to the common traffic flow changes in practice. Therefore, we can conclude that the DRI-MSC model converge fast with low fluctuation in these representative conditions. 

Fig. \ref{fig_9b} shows the average phase duration of the policies. We can see that the average duration of phases for the through lanes is always longer than that of the phases for left-turn directions.  Moreover, the duration of the phases keeps increasing with the increase of traffic flow, since heavier traffic requires longer phase duration. In addition, when the traffic of north-south through direction and north-south left-turn direction is heavier than east-west through direction and east-west left-turn direction, the corresponding signal light duration also changes accordingly. Therefore, the DRI-MSC model is proved to be capable of optimizing the efficiency of traffic conditions. Furthermore, the DRI-MSC policy can also be regarded as a kind of duration optimization policy, since the DRI-MSC policy based phase duration varies according to different traffic flows adaptively.

\begin{figure}[ht]
\centering
\includegraphics[width=1\linewidth]{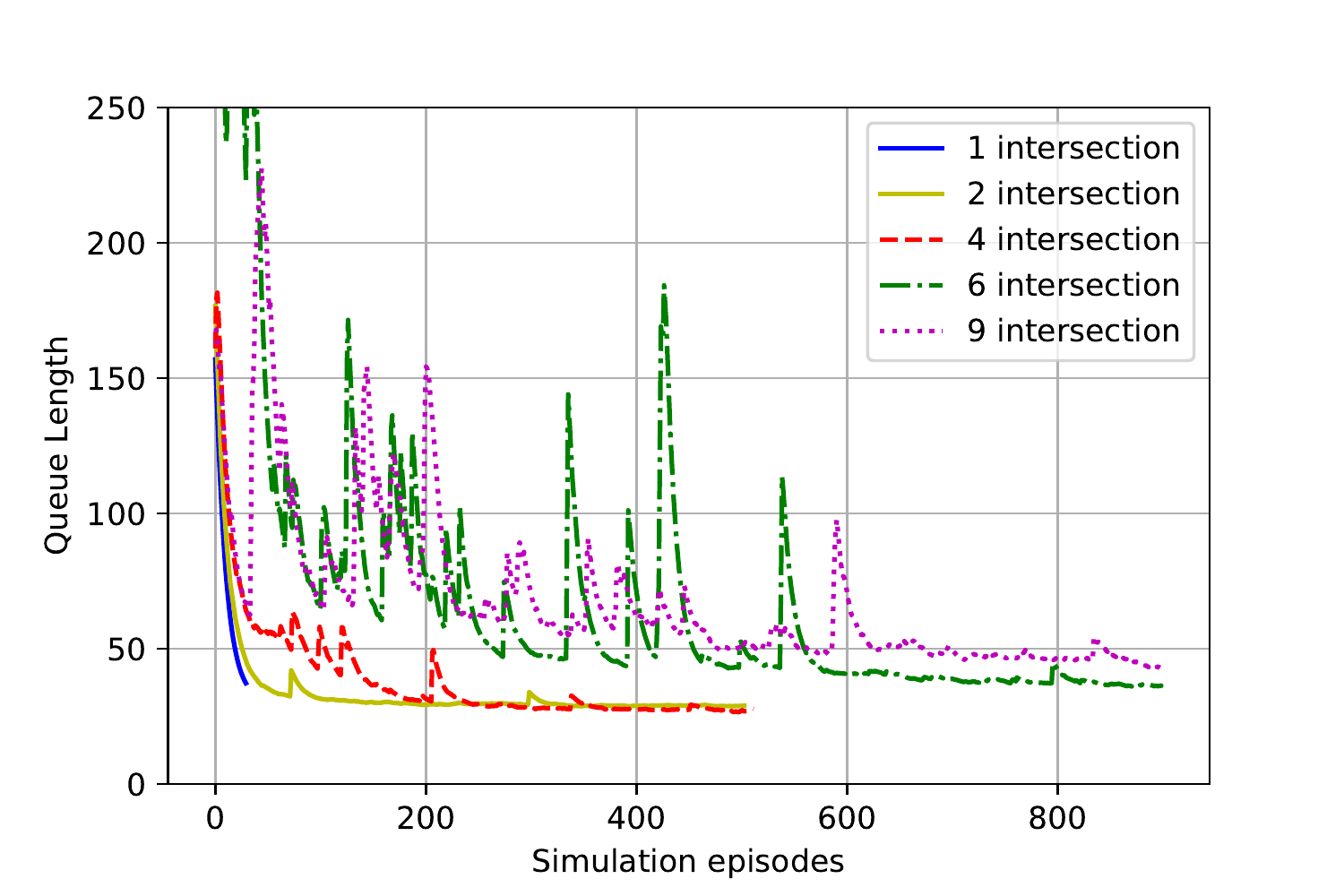}
\caption{Converging curves of DRI-MSC model}
\label{fig12}
\end{figure}

Next, we vary the number of intersections and compare the performance of the DRI-MSC model in Fig. \ref{fig12}. 

Fig. \ref{fig12} shows that the DRI-MSC  is able to converge along with iterations. For the traffic network with no more than 4 intersections, the DRI-MSC model can stably converge to a point with good global efficiency within 200 simulation episodes. Since the action space gradually increases with more intersections, the convergence rate slows down with the expansion of traffic network and more training iterations are needed to explore the model. The DRI-MSC model can also converge with 6 and 9 intersections, if more simulation episodes are reached.

\begin{figure}[ht]
\begin{subfigure}{0.5\textwidth}
\centering
\includegraphics[width=1\linewidth]{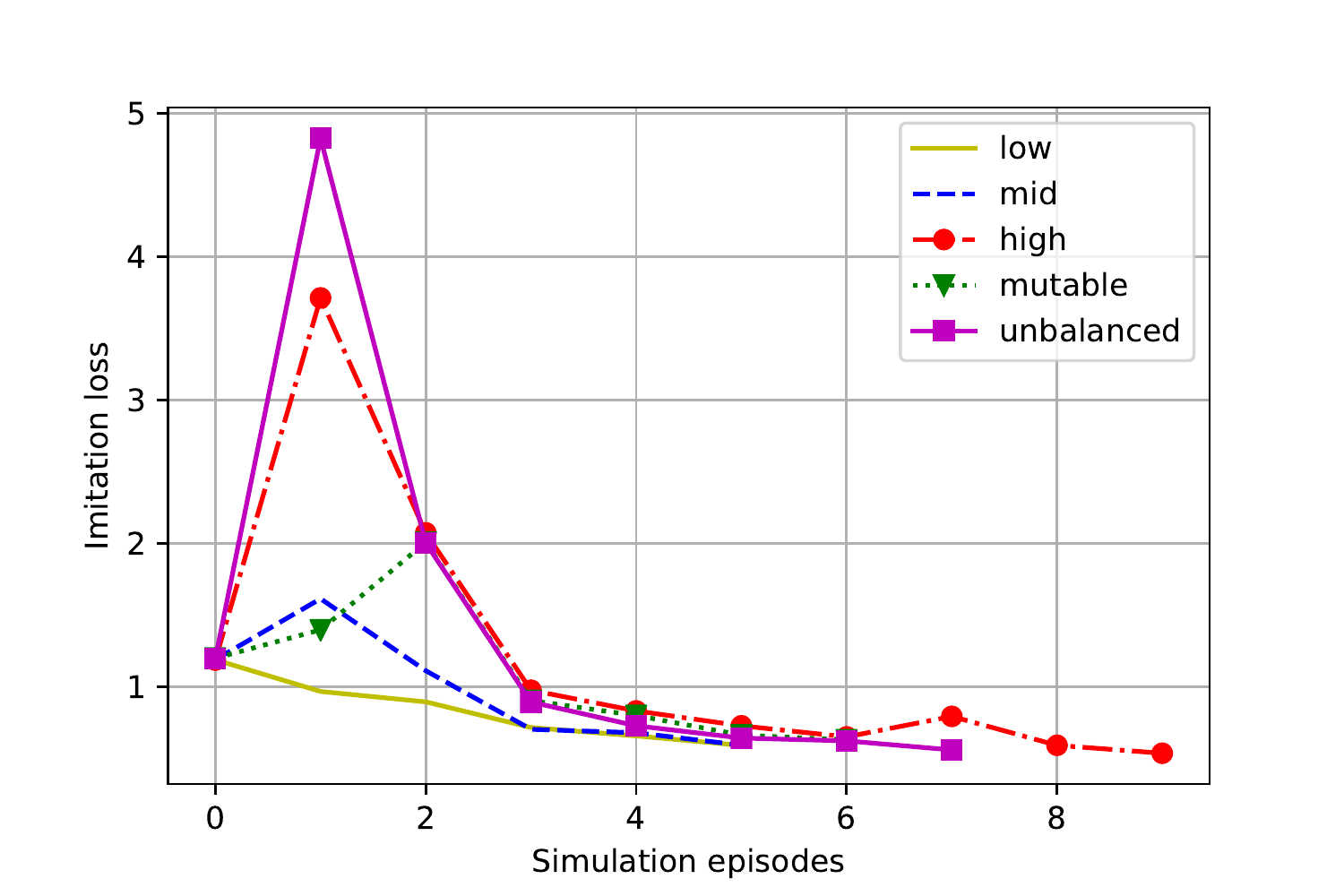}
\caption{Losses of different flows.}
\label{fig_7a}
\end{subfigure}
\begin{subfigure}{0.5\textwidth}
\centering
\includegraphics[width=1\linewidth]{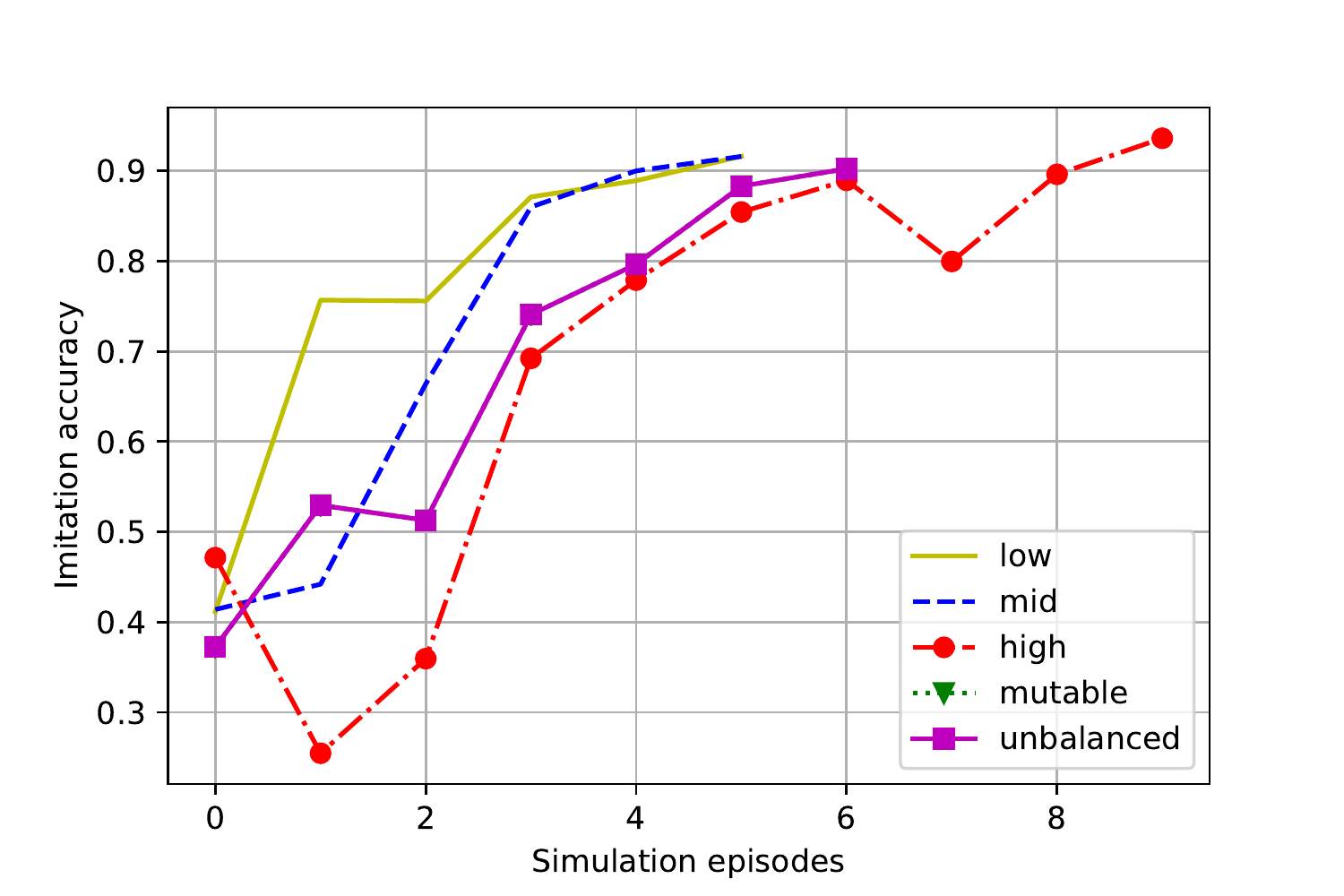}
\caption{Accuracies of different flows.}
\label{fig_7b}
\end{subfigure}
\caption{Performance of imitation learning method of single intersection model with different traffic flows }
\label{fig7}
\end{figure}

Furthermore, to give more insights into the DRI-MSC model, we evaluate the detailed strategies and mechanisms of the DRI-MSC model respectively. Firstly, we evaluate the efficacy of the imitation learning, where different flows with various numbers of intersections are used to test the accuracies brought by imitating the existing policies. Fig. \ref{fig7} describes the variation of errors and accuracies of single intersection model with different traffic flows. 

\begin{figure}[ht]
\begin{subfigure}{0.5\textwidth}
\centering
\includegraphics[width=1\linewidth]{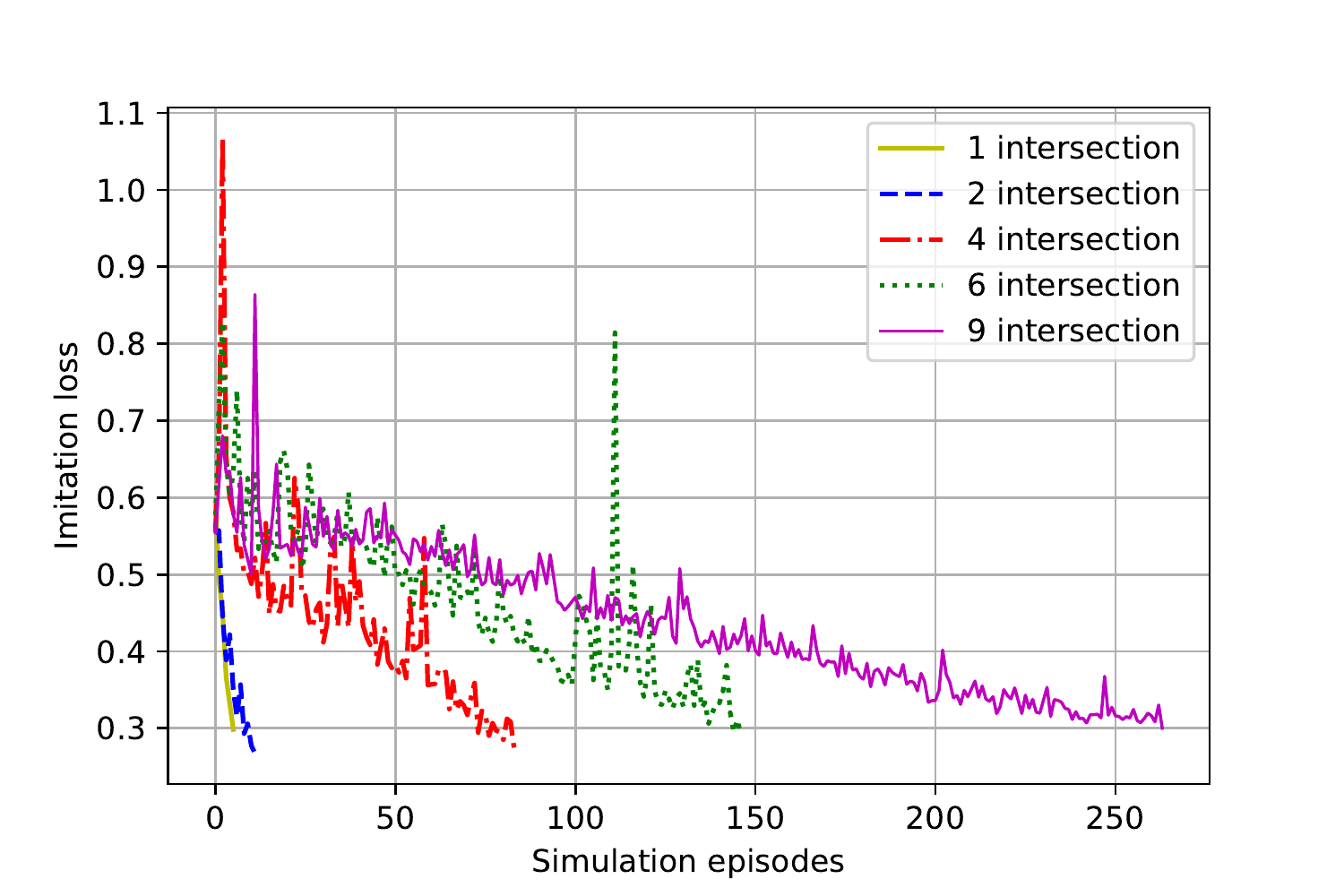}
\caption{Losses of different intersection numbers.}
\label{fig_7c}
\end{subfigure}
\begin{subfigure}{0.5\textwidth}
\centering
\includegraphics[width=1\linewidth]{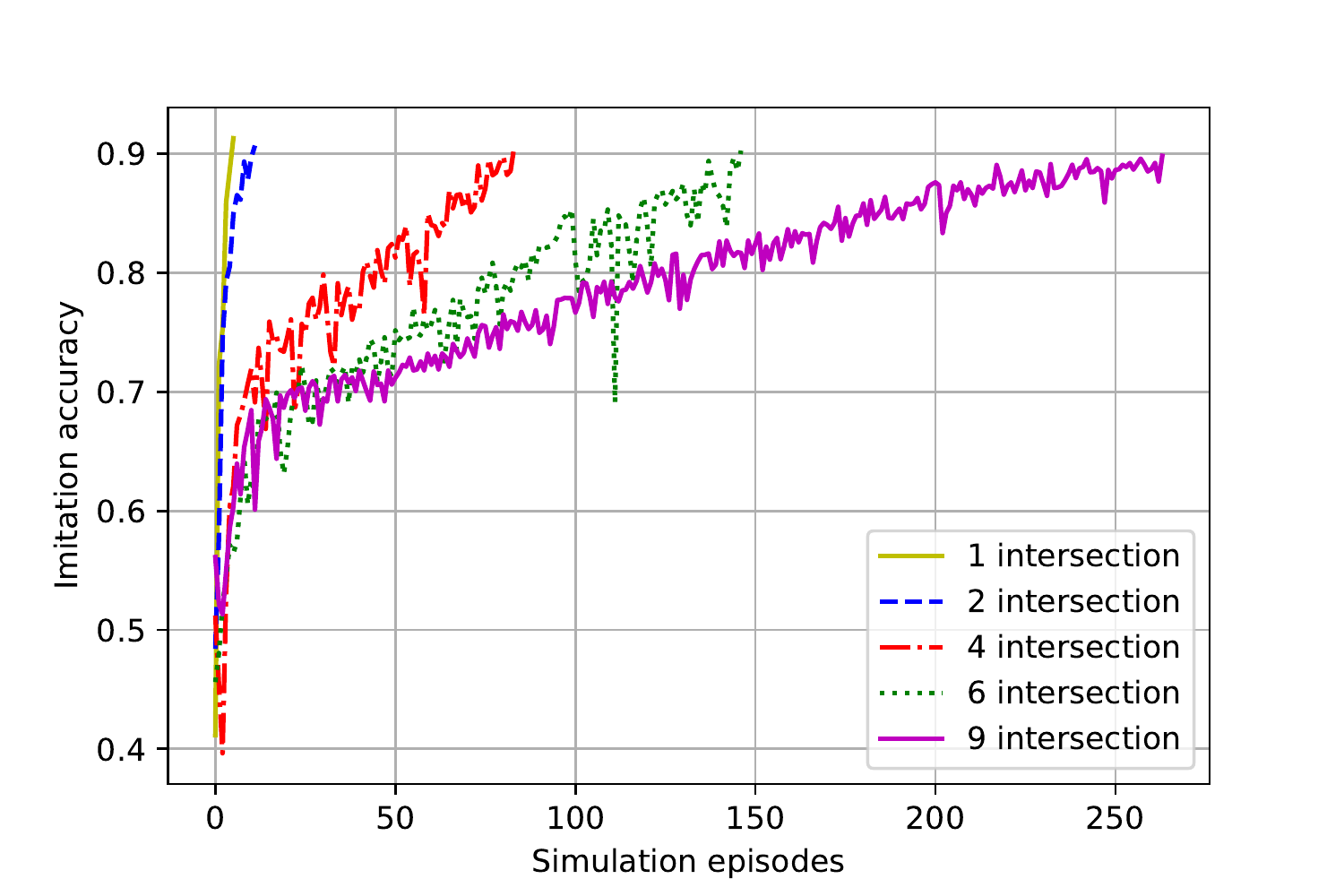}
\caption{Accuracies of different intersection numbers.}
\label{fig_7d}
\end{subfigure}
\caption{Performance of imitation learning method with different number of intersections with traffic flow fixed as low level. }
\label{fig77}
\end{figure}

It can be seen from Fig. \ref{fig7} that the imitation loss decreases and the accuracy increases with simulation episodes going on, and then they both converge to a satisfactory solution. Fig. \ref{fig_7c} and \ref{fig_7d} describe the performance with different numbers of intersections under the same traffic flow level.

Fig \ref{fig77} shows the same tendency with Fig \ref{fig7}. Although the convergence speed becomes slower with more intersections, a satisfactory accuracy can still be ultimately obtained at the end. Therefore, the imitation method  performs well under the situations of various intersections with different traffic flows.

\begin{figure}[t]
\begin{subfigure}{0.5\textwidth}
\centering
\includegraphics[width=1\linewidth]{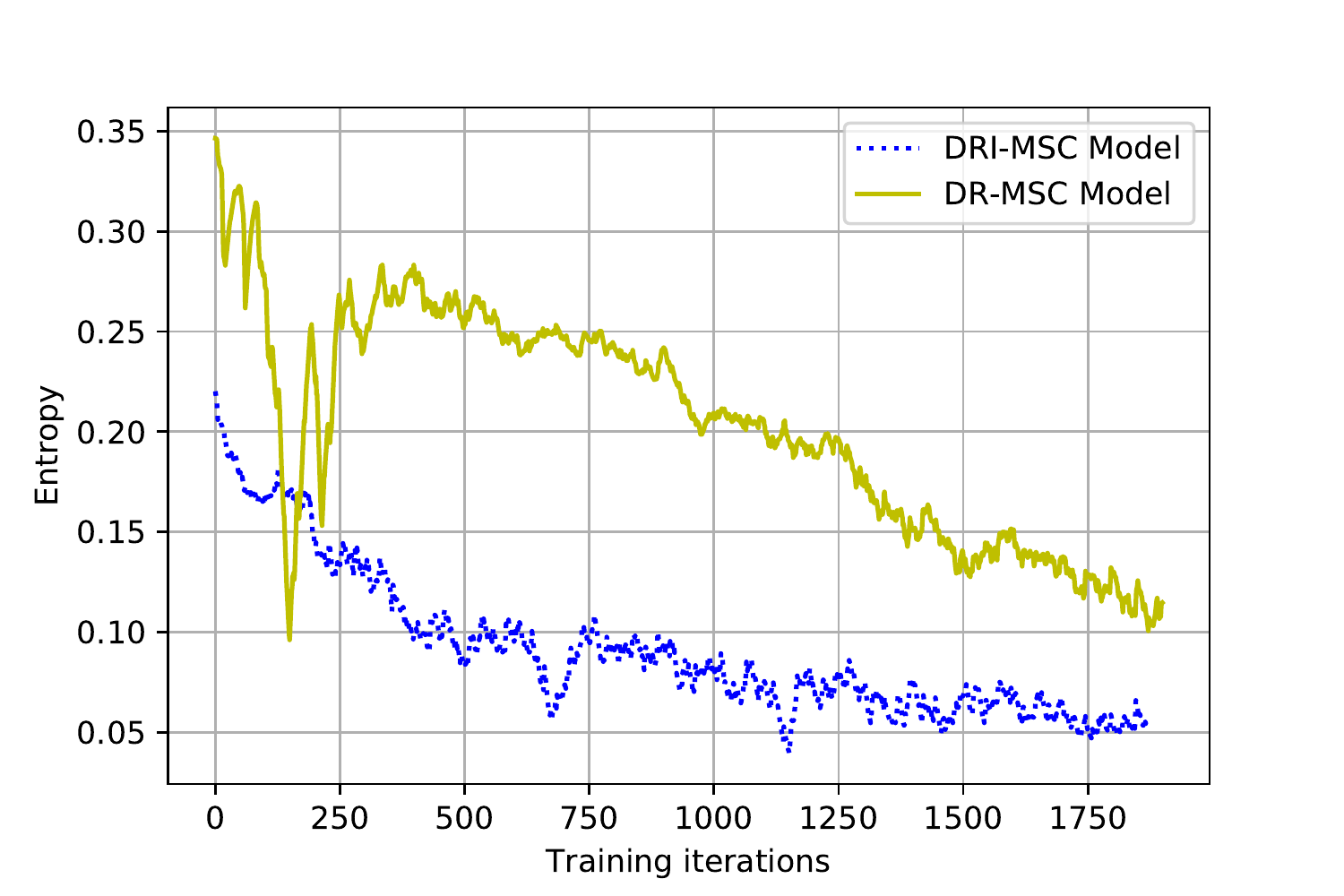}
\caption{Entropy}
\label{fig_8a}
\end{subfigure}
\begin{subfigure}{0.5\textwidth}
\centering
\includegraphics[width=1\linewidth]{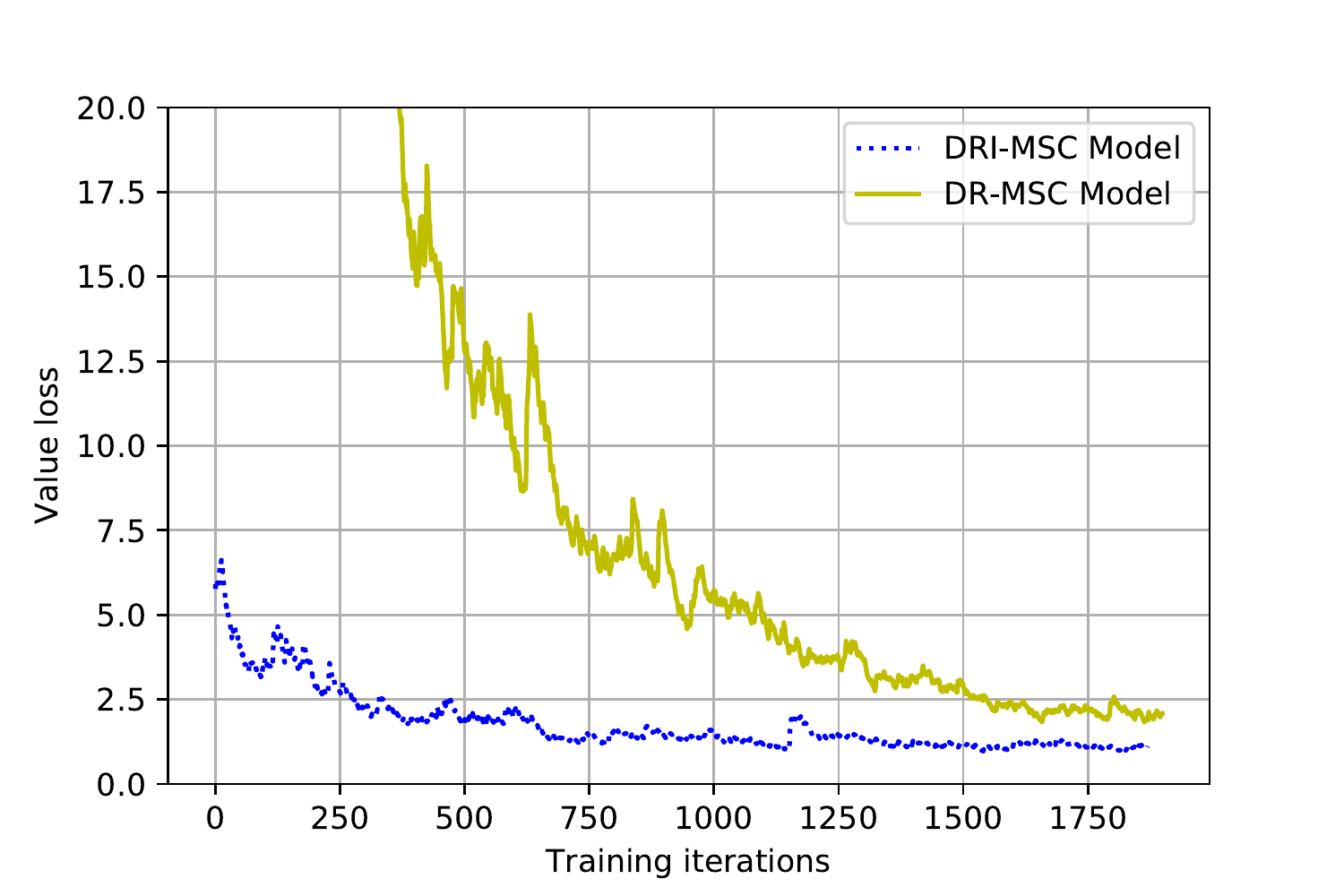}
\caption{Value loss}
\label{fig_8b}
\end{subfigure}
\caption{Comparison between DRI-MSC model and RL based DR-MSC model}
\label{fig8}
\end{figure}

 As shown in Fig. \ref{diff_model},  for the  DRI-MSC model, the left part of the vertical magenta dotted line uses imitation learning as the pre-training, and the right part uses RL for fine-tuning. We can see that the  DRI-MSC model converges significantly faster than the DR-MSC model. It can also be noticed that the converging process of the DRI-MSC model is more stable with less fluctuation. For further analysis, we adopt entropy shown in Fig \ref{fig_8a} to quantify the convergence of the probability based control policies. It is obvious that the entropy of the DRI-MSC model is distinctively lower than that of the DR-MSC model, meaning that the policy already converges to a good point via the pre-training. Therefore, imitation learning can help the model skip many redundant searches.

For the RL fine training, actor critic method is used, in which the policies and values converge in complementary to avoid fluctuation.  We also compare the convergence of value loss, which is demonstrated in Fig. \ref{fig_8b}. Although  the value loss is not taken into account in the pre-training, it can also be reduced by imitation learning.  This is mainly due to the fact that the actor and critic share parameters, and the gradients of value and policy move roughly in the same direction.  Therefore, both the policy and value benefit from the pre-training.

\begin{figure}[ht]
\centering
\includegraphics[width=1\linewidth]{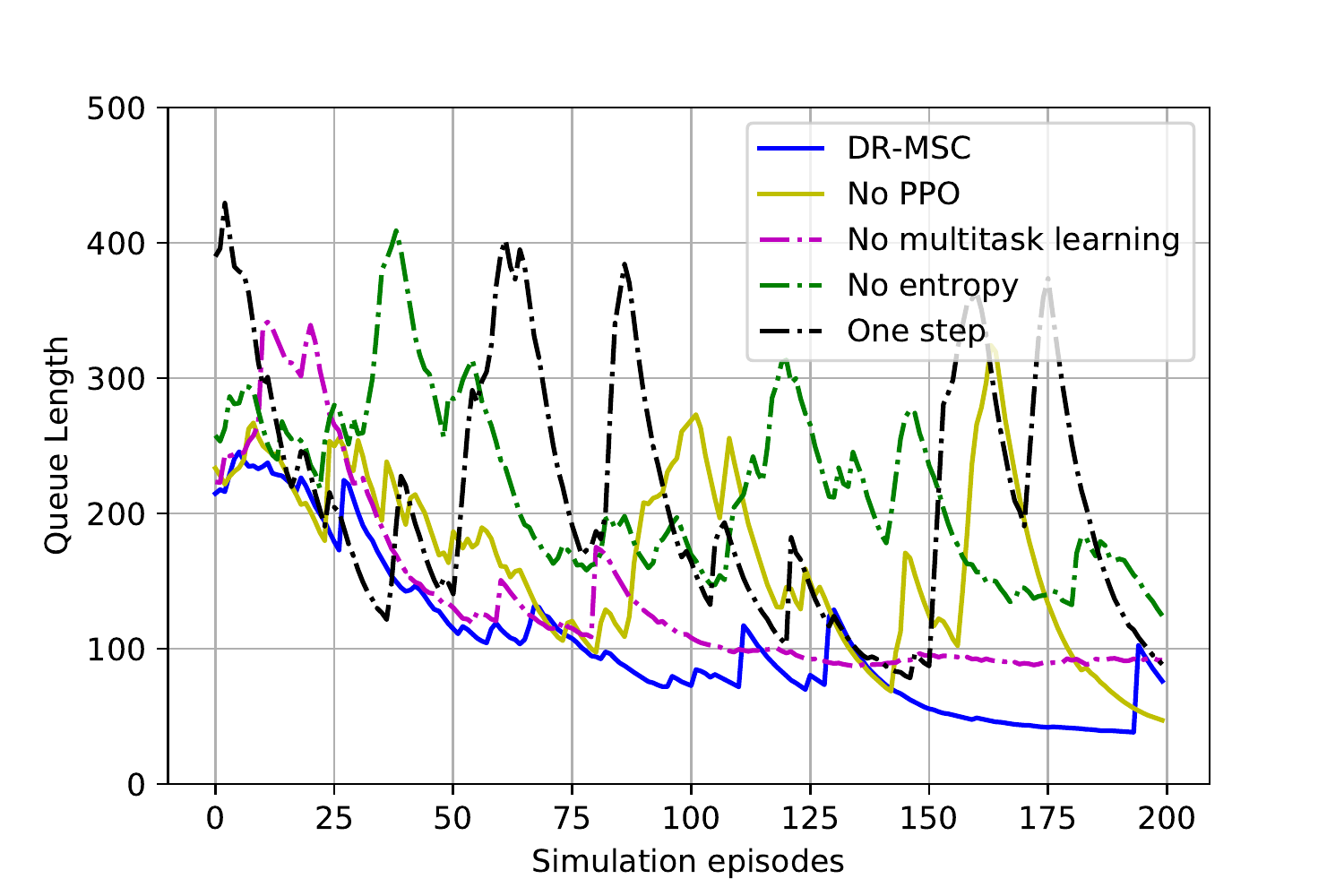}
\caption{Comparison of different boosting methods}
\label{fig10}
\end{figure}

For methods of boosting the training for multi-intersections, we evaluate the effects of multi-step estimation, multi-task learning and PPO method in the overall performance of DR-MSC model respectively in Fig. \ref{fig10}. For the model without PPO, we use simple policy gradient instead. For the model without multi-task learning, we use one-task output, in which we use different neural networks to output the actions of the four intersections and the critic value $v$. For the model without soft RL, we add an $l_2$ penalty with coefficient 0.01.

In Fig. \ref{fig10}, the DR-MSC model works best, and PPO method can significantly improve the convergence of the model. Moreover,  we can see that the DR-MSC model converges slightly faster than the model without multi-task learning method, whose overall efficiency can not be lower than 80m either. This is because that multi-task learning method can learn a more general representation for all the tasks to avoid overfitting, and reduce the task-dependent noise which negatively affect the overall performance. Finally, the model with $l_2$ regularization converges much slower than  DR-MSC model, since the soft RL method penalizes the policy directly which is more effective compared with the penalty to the parameter space.

Lastly,  Table \ref{efficiency_compare} is given to further test the efficiency of the  proposed model under different flow conditions of isolated intersection, where ``QL", ``AWT", ``AFC", "Rule", "IL" and "Unb" represents queue length, average vehicle waiting time, average vehicle fuel consumption, rule-based policy defined in (\ref{rule}), and unbalanced flow respectively. Two popular fixed timing policies  are also compared.

\begin{table}[ht]
\caption{Performance of different models}\label{efficiency_compare}
\begin{tabular}{ m{0.5cm} m{1.8cm} m{0.7cm}  m{0.7cm} m{0.7cm} m{0.7cm}  m{0.8cm}<{\centering}  }

\Xhline{1.2pt}
     & Flow & Low  &Middle  & High & Mutable  &Unb \\

\hline
\multirow{6}{0.5cm}{QL } &   Fixed-time 20   &75.30 &115.30& 849.21&153.77 &621.01
\\
 &  Fixed-time 40  &134.62 &202.48 &810.95 &183.34 &597.60
\\
 &Rule &36.83 &102.39 &442.23 &105.69 &103.22
\\
&IL  &36.50 &108.35 &453.27 &114.39 &104.88
\\
&DR-MSC &\textbf{30.60} &93.39 &362.63 &102.23 &109.61
\\
 &DRI-MSC &30.90 &\textbf{86.91} &\textbf{342.23 }&\textbf{101.54} &\textbf{85.36}
\\
\hline
\multirow{6}{0.5cm}{AWT } &  Fixed-time 20  &18.09 &18.42&52.00&21.59&36.23
\\
 &Fixed-time 40 & 33.43 &32.80 &48.58 &32.91 &47.39
\\
 &Rule  & 8.84&14.97 &37.65&14.16 &17.18
\\
&IL &8.69&14.77 &38.85&15.50 &16.67
\\
&DR-MSC  &\textbf{7.61}&13.07 &36.00&12.22 &15.73
\\
 &DRI-MSC  & 7.62 &\textbf{12.59}&\textbf{34.66}&\textbf{11.68 }&\textbf{11.98}
\\
\hline
\multirow{6}{0.5cm}{AFC} &  Fixed-time 20  &55.23 &78.27&244.05&83.69&178.00
\\
 &Fixed-time 40 & 69.62 &99.27 &245.57 &90.79 &176.57
\\
 &Rule  & 47.01&75.47 &174.81&69.79&71.28
\\
&IL  &46.64&77.22 &176.75&74.21&71.98
\\
&DR-MSC  &45.96&72.86&157.08&71.17 &73.15
\\
 &DRI-MSC &\textbf{\textbf{45.77}} &\textbf{71.82}&\textbf{152.63}&\textbf{71.08}&\textbf{66.83}
\\

\Xhline{1.2pt}
\end{tabular}
\end{table}

Table \ref{efficiency_compare} demonstrates that the fixed timing policy performs well only under a few certain conditions. In contrast, the DRI-MSC model shows to be advantageous in various environments, especially under the condition of large flow and unbalanced flow. We can also see that the imitation learning can achieve similar results as the rule based. The DR-MSC model performs well in all the conditions and goes beyond the rule based model, meaning that the data driven RL method can even achieve a level surpassing the human design.  Generally, the DRI-MSC model performs best in all the 3 indexes given. 

 \section{Conclusion}\label{sect4}
 This paper propose a novel end-to-end model based on deep imitation learning and reinforcement learning for multi-intersection signal control. Instead of using a huge matrix, the proposed model introduce tensors to extract information as the inputs, and multi-dimensional boolean output is also used to simplify the representation without information loss. The multi-task learning is also introduced in the model.  In the training, the proposed model achieves significantly fast convergence by using imitation learning as the pre-training with proximal policy optimization. Experiment results show the efficiency and accuracy of the proposed model with feasibility for real-world applications. 

For future work, incomplete information environment can be focused due to the low equipment permeability in the real world. Moreover,  parallel learning can be introduced to combine the off-line learning and on-line learning, and transfer reinforcement learning can also be potentially used to handle more task-specific  scenarios.

\appendices

\ifCLASSOPTIONcaptionsoff
  \newpage
\fi

\bibliographystyle{IEEEtran}
\bibliography{ref}

\begin{IEEEbiography}[{\includegraphics[width=1in,height=1.25in,clip,keepaspectratio]{{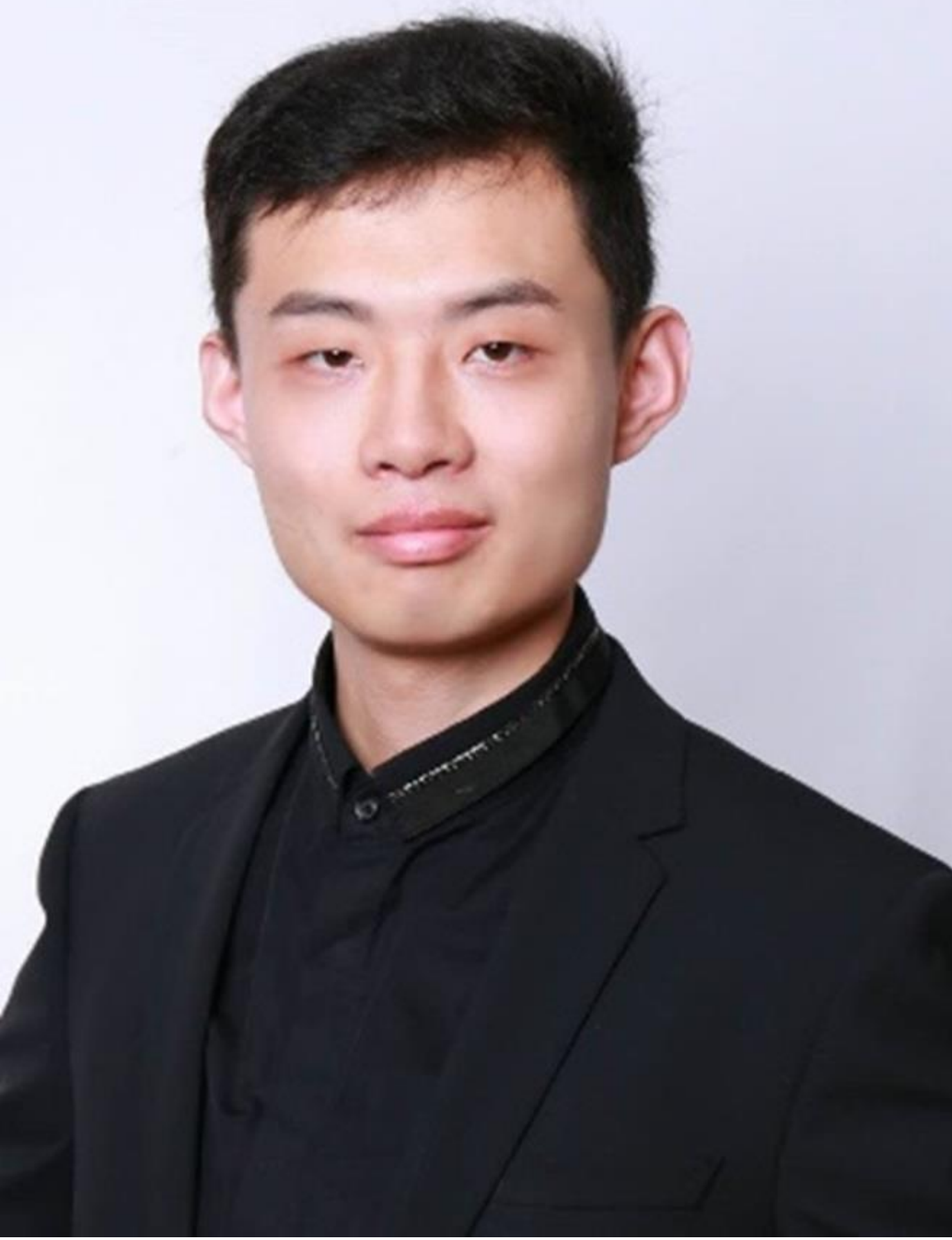}}}]{Yusen Huo}
is currently a graduate student in Tsinghua University. His research interests include intelligent transportation systems, deep
learning and reinforcement learning.
\end{IEEEbiography}

\begin{IEEEbiography}
[{\includegraphics[width=1in,height=1.25in,clip,keepaspectratio]{{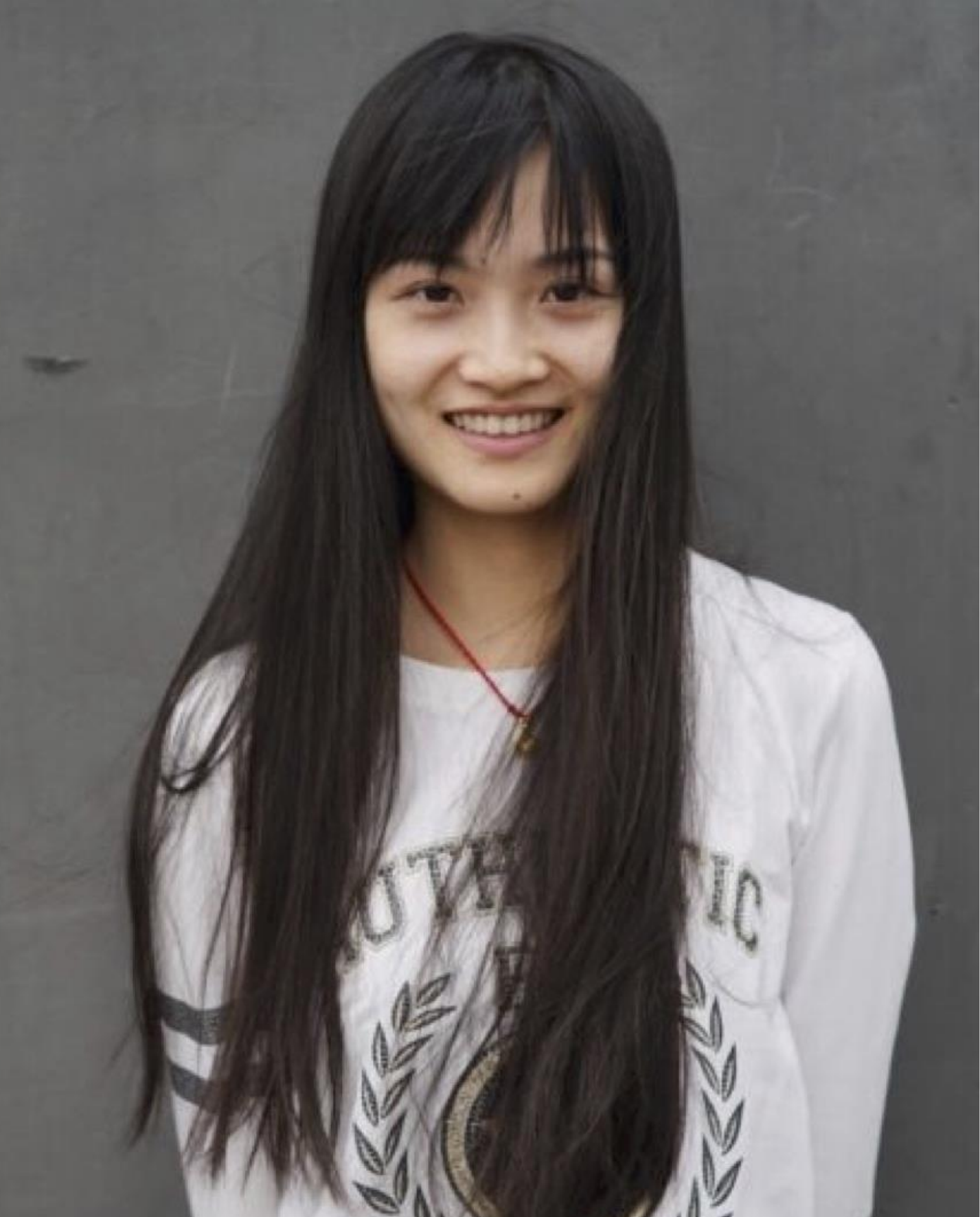}}}]
{Qinghua Tao}
received her B.S. degree from Central South University, Changsha, China, in 2014. She is currently pursuing the Ph.D. degree in the Department of Automation, Tsinghua University, Beijing, China. From 2017 to 2018, she was a visiting Ph.D student in ESAT, KU Leuven, Belgium. Her research interests include the modeling and training of piecewise neural networks, the sparsity regularization with piecewise linearity, derivative-free optimization for black-box problem, and other relating machine learning methods.    

\end{IEEEbiography}

\begin{IEEEbiography}[{\includegraphics[width=1in,height=1.25in,clip,keepaspectratio]{{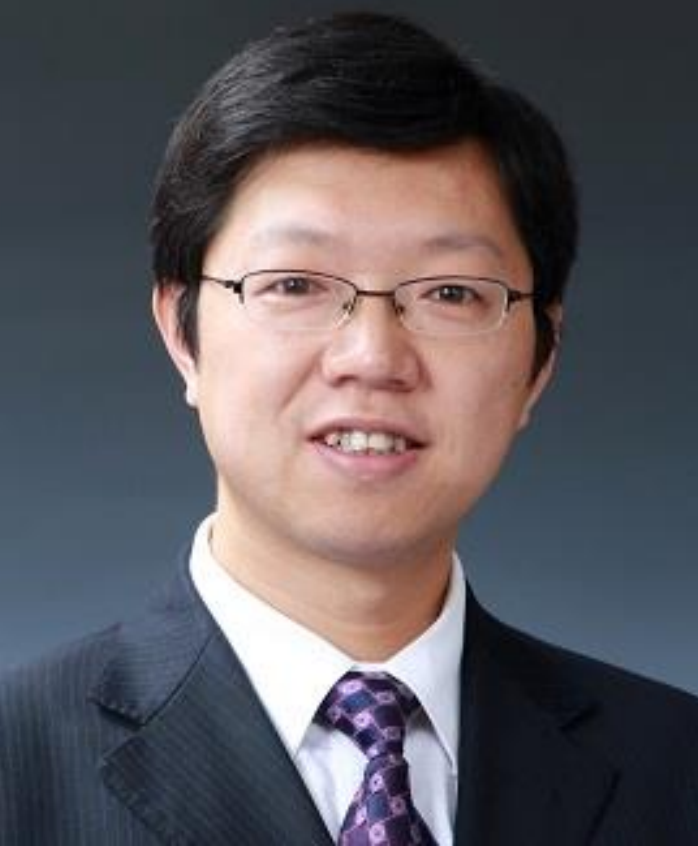}}}]{Jianming Hu}
is currently an Associate Professor of Department of Automation (DA), Tsinghua University. He got his B.E., M.E. and Ph.D.’s degrees in 1995, 1998 and 2001 respectively. He worked in Chinese University of Hong Kong from 2004 to 2005 as a research assistant and visited PATH, University of California at Berkeley for one year as a visiting scholar. He has presided and participated over 20 research projects granted from Ministry of Science and Technology of China, National Science Foundation of China and other large companies with over 30 journal papers and over 90 conference papers. Based on the research achievements, he obtained two Sci. $\&$ Tech. Improvement Awards from Ministry of Education and ITS Association of China in 2016 and 2012 respectively, the 1st class Award of Excellent Teaching Skill Contest in Beijing and many awards from Tsinghua University. Dr. Hu’s recent research interests include networked traffic flow, large scale traffic information processing, intelligent vehicle infrastructure cooperation systems (V2X or Connected Vehicles), and urban traffic signal control etc. 
\end{IEEEbiography}

\enlargethispage{-5in}

\end{document}